\documentclass[reprint,aps,pre,showpacs,superscriptaddress]{revtex4-2}

\usepackage{graphicx}
\usepackage{dcolumn}
\usepackage{bm}
\usepackage{bbm} 
\usepackage{amsfonts} 
\usepackage{url}

\usepackage[ruled,vlined]{algorithm2e}
\usepackage{hyphenat}
\usepackage{graphicx}
\usepackage{natbib}
\usepackage{caption}
\usepackage{siunitx}

\newcommand{\pdv}[2]{\frac{\partial #1}{\partial #2}}
\usepackage{xparse}
\RenewDocumentCommand{\pdv}{ o m m g }{%
  \IfNoValueTF{#1}
    {
      \IfNoValueTF{#4}
        {\frac{\partial #2}{\partial #3}} 
        {\frac{\partial^2 #2}{\partial #3 \partial #4}} 
    }%
    {
      \IfNoValueTF{#4}
        {\frac{\partial^{#1} #2}{\partial #3^{#1}}} 
        {\frac{\partial^{#1} #2}{\partial #3^{#1-1}\partial #4}} 
    }%
}
\usepackage{siunitx}
\usepackage{nicefrac}
\usepackage{commath}
\usepackage{multirow}
\usepackage{makecell}
\usepackage{booktabs}
\usepackage{nicefrac}
\usepackage[inline, shortlabels]{enumitem}
\usepackage{siunitx}
\usepackage{caption}

\begin{document}

\preprint{APS/123-QED}

\title{\textbf{Physics-Informed Neural Networks for Solving Derivative-Constrained PDEs}}
\author{Kentaro Hoshisashi}
 \email{Contact author: k.hoshisashi@ucl.ac.uk}
\author{Carolyn E. Phelan}
\author{Paolo Barucca}
\affiliation{
Department of Computer Science,\\ University College London,\\ Gower Street, London WC1E 6BT, UK
}

\date{\today}

\begin{abstract}
  Physics-Informed Neural Networks (PINNs) recast PDE solving as an optimisation problem in function space by minimising a residual-based objective, yet many applications require additional derivative-based relations that are just as fundamental as the governing equations. In this paper, we present Derivative-Constrained PINNs (DC-PINNs), a general framework that treats constrained PDE solving as an optimisation guided by a minimum objective function criterion where the physics resides in the minimum principle. DC-PINNs embed general nonlinear constraints on states and derivatives, e.g., bounds, monotonicity, convexity, incompressibility, computed efficiently via automatic differentiation, and they employ self-adaptive loss balancing to tune the influence of each objective, reducing reliance on manual hyperparameters and problem-specific architectures. DC-PINNs consistently reduce constraint violations and improve physical fidelity versus baseline PINN variants, representative hard-constraint formulations on benchmarks, including heat diffusion with bounds, financial volatilities with arbitrage-free, and fluid flow with vortices shed. Explicitly encoding derivative constraints stabilises training and steers optimisation toward physically admissible minima even when the PDE residual alone is small, providing reliable solutions of constrained PDEs grounded in energy minimum principles.
\end{abstract}

\maketitle

\section{Introduction}
Physics-informed neural networks (PINNs) \cite{raissi2019physics} have recently emerged as a powerful deep learning framework to solve partial differential equations (PDEs). By embedding PDE residuals directly into the loss function of a neural network, PINNs recast the task of solving PDEs as an optimisation problem in the function space, thus shifting the focus from discrete numerical schemes to a variational viewpoint that naturally connects to broader physical principles \citep{sirignano2018dgm, berg2018unified}. Unlike finite-difference and finite-element solvers, which remain highly effective but can become costly in many-query settings such as inverse problems, parameter inference, and uncertainty quantification, as well as on complex or evolving geometries that require repeated remeshing, PINNs provide a flexible formulation that can incorporate additional physical structure.

This reformulation represents a theoretical shift from viewing the PDE as a forward operator to interpreting it as the condition that the system minimises a variational or energy principle \cite{yu2018deep}. 
The physics is encoded not only in the PDE but in the fact that the admissible solution minimises an appropriate functional \cite{kharazmi2021hp}, leveraging a minimum objective function criterion. This perspective has been discussed in a broader scientific context, from minimum dissipation principles in fluid dynamics \cite{onsager1931reciprocal, patel2024turbulence}, to energy-based formulations in statistical mechanics and quantum systems \citep{karniadakis2021physics}, and to physics-guided optimisation principles in complex systems \citep{chen2021physics}. In this perspective, PINNs and their extensions provide a computational framework that leverages this minimum principle by embedding the governing equations and constraints into a single objective, and training the network corresponds to searching for the physical extremum. Such constraints can be as fundamental to physics as the PDE itself, and neglecting them risks producing solutions that are mathematically feasible but physically implausible \cite{karniadakis2021physics}.

However, despite their empirical successes, PINNs still face challenges in the presence of additional equality or inequality constraints beyond the PDE itself, as pioneered in \cite{lagaris1998artificial, krishnapriyan2021characterizing, cuomo2022scientific, wang2022respecting}. 
Such constrained PDEs are ubiquitous in real-world applications, e.g., bounds, monotonicity, convexity, incompressibility, to name a few. For instance, incompressible flow models rely on derivative consistency to link pressure gradients and velocity fields, and violations can lead to non-physical wake dynamics.
In quantitative finance, no-arbitrage requirements translate into derivative inequalities (e.g., monotonicity and convexity) that render a calibrated surface unusable if violated.
Even when derivative constraints are not explicitly imposed, the adaptive balancing component introduced here is designed to reduce sensitivity to manual loss-weight tuning, although we only validate it in the derivative-constrained setting in this manuscript.
Naive techniques for embedding constraints can lead to unstable training, slow convergence, and violation of constraints \cite{wang2022and}. Although more sophisticated approaches have been proposed to better handle constraints in PINNs, conservative PINNs, and enforce equality constraints, while augmented Lagrangian approaches and theory-guided neural networks show stronger performance in PDEs constrained by inequality~\citep{jagtap2020conservative, lo2023training, sukumar2022exact, lu2021physics, chen2021theory, beucler2021enforcing}.
However, a persistent difficulty lies in balancing competing error terms, the residual PDE, boundary conditions, and constraint penalties, which often differ in magnitude and sensitivity. As a result, these methods remain highly dependent on delicate hyperparameter tuning and problem-specific architectural choices.

In this study, we propose Derivative-Constrained PINNs (DC-PINNs), a general and robust framework for solving derivative-constrained PDEs using deep learning. Our approach seamlessly integrates the constraints into the learning process, ensuring that the solution satisfies the prescribed conditions while maintaining the benefits of PINNs. The key contributions of this study include:

\vspace{-10pt}\begin{itemize}
    \item A flexible constraint-aware loss function that admits general nonlinear constraints and seamlessly incorporates them into the PDE residual objective.
    \vspace{-5pt}\item Self-adaptive loss balancing techniques that automatically tune the weightings of the objective terms, including derivatives obtained through automatic differentiation, stabilising training across diverse problem settings.
    \vspace{-5pt}\item Demonstration of the approximation ability of DC-PINNs on benchmark PDEs from simple to complex, including the heat equations, structural modelling of finance, and fluid dynamics problems, showcasing improvements over PINNs approaches.
\end{itemize}

The study aims to highlight the importance of explicit constraint handling in PINNs, as relying solely on the PDE residual term may lead to solutions that violate critical physical principles. 
The extended approach can explore an application to several key areas:
\begin{enumerate*}[series = tobecont, itemjoin = \quad]
   \item Basic PDE problems with added physical insights
   \item Problems requiring multiple derivative constraints to avoid system failure
   \item Complex problems with diverse internal dynamics that typically resist convergence.
\end{enumerate*}
We present applications including heat equations, volatility surface models in finance, and incompressible flows in fluid dynamics, comparing our framework with existing approaches. Recasting derivative-constrained PDE solving as a multi-objective optimisation problem opens up the possibility of efficiently solving a wider range of physically constrained systems with improved accuracy and reliability.

\paragraph*{Baselines and fairness.}
Standard PINNs do not encode derivative inequalities. We therefore include (i) \textit{PINNs+ineq(fixed)} which adds derivative inequality attentions with fixed global weights, and (ii) hard-constraint baselines (\textit{hPINNs (pen.)}, \textit{hPINNs (AL)}, \textit{AL-PINNs}) in which inequalities are treated by homotopy or augmented Lagrangians. Compared to these, DC-PINNs uses one-sided penalties on derivative constraints but learns both category-level coefficients and sample-level scalings, which we ablate to disentangle the benefit of constraints vs. adaptivity.

\section{Problem Formulation}
\subsection{Derivative-Constrained PDE Problem}
\label{sec:constrained-pde}
We seek a mapping $u:\Omega\subset\mathbb{R}^n\to\mathbb{R}^d$ that is approximated by a neural network $u_\theta\equiv\phi_\theta(\boldsymbol{x})$.  
Let $\mathcal{D}$ denote the set of differential operators applied to $u_\theta$, including the zeroth-order identity. Training is posed as a constrained optimisation problem.
\begin{equation}
\hat\theta=\arg\min_{\theta}\,\mathcal{L}\big(\boldsymbol{x},\mathcal{D}u_\theta\big)
\,\,\,\text{s.t.}\,\,
\begin{cases}
f\big(\boldsymbol{x},\mathcal{D}u_\theta\big)=0 & \boldsymbol{x}\in\Omega,\\[2pt]
b\big(\boldsymbol{x},\mathcal{D}u_\theta\big)=0 & \boldsymbol{x}\in\partial\Omega,\\[2pt]
h\big(\boldsymbol{x},\mathcal{D}_h u_\theta\big)\le 0 & \boldsymbol{x}\in\Omega,
\end{cases}
\label{eq:constrainedPDE}
\end{equation}
where $\mathcal{L}$ is the training objective, $f$ represents PDE residuals in the domain, $b$ represents boundary operator and $h$ represents derivative inequality constraints.  
Here, $\mathcal{D}_h\subseteq\mathcal{D}$ specifies the derivatives that appear in the inequalities. A problem is called \emph{derivative-constrained} when it includes inequality conditions on derivatives of $u$. Typical examples are monotonicity constraints ($\nabla u\ge 0$), directional convexity constraints ($\mathrm{diag}(\nabla^2 u) \ge 0$), or slope bounds ($\|\nabla u\|\le L$). This notion refers to constraints on derivatives and not to the order of the PDE itself, and reflects physical phenomena that cannot be fully expressed by the PDE alone.

For discretisation, we employ supervised data $D_0=\{(\boldsymbol{x}_0,u_0)\}=\{(\boldsymbol{x}_i^{(0)},u_i^{(0)})\}_{i=1}^{N_0}$, boundary points $\boldsymbol{x}_b=\{\boldsymbol{x}_i^{(b)}\}_{i=1}^{N_b}$, interior collocation points $\boldsymbol{x}_f=\{\boldsymbol{x}_i^{(f)}\}_{i=1}^{N_f}$, and inequality-enforcement points $\boldsymbol{x}_h=\{\boldsymbol{x}_i^{(h)}\}_{i=1}^{N_h}$.

This work focusses on inequality constraints involving derivatives and adopts a discretise, then optimise strategy with gradient-based training of artificial neural networks (ANNs). The continuous problem is first discretised into a finite-dimensional nonlinear problem. The derivatives with respect to the inputs are evaluated by automatic differentiation (AD) through the typical ANN, that is, the multilayer perceptron (MLP). To ensure that all required partial derivatives exist and are well defined, the activation function must be sufficiently smooth with respect to the inputs. For PDEs that involve derivatives up to order~$k$, the network output must be at least $C^{k}$ continuous. Because training employs gradient-based optimisation through these derivatives, the activation should in practice be $C^{k+1}$, which is satisfied, for example, by the hyperbolic tangent ($\tanh$) activation function.

\subsection{Physics-Informed Neural Networks (PINNs)}
Physics-Informed Neural Networks (PINNs), introduced in~\cite{raissi2019physics}, are artificial neural networks that integrate physical laws directly into the training objective via partial differential equations (PDEs). This paradigm offers a data-efficient class of universal function approximators in which the governing PDE residuals and boundary or initial conditions are treated as components of the loss function.  

We consider a parametrised PDE system of the form
\begin{equation}
\begin{cases}
f\bigl(\boldsymbol{x},\mathcal{D}u_\theta(\boldsymbol{x})\bigr)=0, & \boldsymbol{x}\in\Omega, \\[2pt]
b\bigl(\boldsymbol{x},\mathcal{D}u_\theta(\boldsymbol{x})\bigr)=0, & \boldsymbol{x}\in\partial\Omega,
\end{cases}
\end{equation}
where $u_\theta\equiv \phi_\theta(\boldsymbol{x})$ denotes the neural network approximation of the true solution $u$. Here, $\mathcal{D}$ denotes the set of differential operators obtained by automatic differentiation (AD) of the neural network with respect to its inputs.

The training problem is formulated as the unconstrained minimisation of a total loss functional,
\begin{equation}
    \mathcal{L} := \mathcal{L}_0 + \mathcal{L}_f + \mathcal{L}_b,
\end{equation}
where the three terms correspond to supervised data loss, residual PDE loss in the domain, and loss of the boundary or initial condition, respectively. Explicitly, one has
\begin{equation}
\begin{gathered}
    \mathcal{L}_0 := \tfrac{1}{N_0}\,\left\|\,u_\theta(\boldsymbol{x}_0)-u_0\,\right\|_2^2, \\
    \mathcal{L}_f := \tfrac{1}{N_f}\,\left\|\,f\bigl(\boldsymbol{x}_f,\mathcal{D}u_\theta(\boldsymbol{x}_f)\bigr)\,\right\|_2^2, \\
    \mathcal{L}_b := \tfrac{1}{N_b}\,\left\|\,b\bigl(\boldsymbol{x}_b,\mathcal{D}u_\theta(\boldsymbol{x}_b)\bigr)\,\right\|_2^2.
\end{gathered}
\end{equation}
Here, $N_0,N_f,N_b$ denote the cardinalities of the supervised dataset, the set of interior collocation points, and the set of boundary collocation points, respectively. Norms are evaluated over the entire vector of residuals at the respective collocation points. The observed targets $u_0$ contribute only to $\mathcal{L}_0$, while $\mathcal{L}_f$ and $\mathcal{L}_b$ depend solely on the PDE operator $f$ and the boundary operator $b$.

\subsection{Derivative-Constrained PINNs (DC-PINNs)}
\label{sec:dc-pinns}
We now extend the preceding framework to the case of derivative-constrained PDEs, as formulated in~\eqref{eq:constrainedPDE}. Specifically, we consider inequality constraints on derivatives of the form
\begin{equation}
    h\bigl(\boldsymbol{x},\mathcal{D}_h u_\theta(\boldsymbol{x})\bigr) \le 0,
    \qquad \boldsymbol{x}\in\Omega,
\end{equation}
where $\mathcal{D}_h\subseteq\mathcal{D}$ denotes the subset of derivatives that appear in the inequality conditions.

In order to enforce such inequality constraints within a gradient-based training scheme, we formulate a one-sided penalty functional. For each inequality operator $h$, we define the violation loss as
\begin{equation}
    \mathcal{L}_{h} := \tfrac{1}{N_h}\,\left\|\,[\,h\bigl(\boldsymbol{x}_h,\mathcal{D}_h u_\theta(\boldsymbol{x}_h)\bigr) ]_+ \,\right\|_2^2,
    \label{eq:inequality_loss_hinge}
\end{equation}
where $[\cdot]_+:=\max\{0,\cdot\}$ denotes the standard hinge operator.
It yields zero gradients inside the feasible region and a linear signal once violations occur, which we found to be numerically stable when combined with AD on input derivatives. Near $h=0$ we smooth the derivative at the origin (C$^1$) to avoid optimiser jitter.
We also tested smooth hinges $\operatorname{softplus}_\delta(h)=\delta\log(1+e^{h/\delta})$ and squared hinges; while comparable in accuracy, they required tighter tuning of global weights to prevent slow drift along nearly-feasible manifolds. The hinge thus strikes a balance between strictness and simplicity for derivative inequalities.

The total loss in the DC-PINN framework generalises the standard PINN loss by incorporating both inequality constraints and weighting coefficients:
\begin{equation}
    \mathcal{L} := 
      \lambda_0 \hat{\mathcal{L}}_0 
    + \lambda_f \hat{\mathcal{L}}_f 
    + \lambda_b \hat{\mathcal{L}}_b 
    + \lambda_h \hat{\mathcal{L}}_{h},
    \label{eq:dc-pinns-loss}
\end{equation}
where $\lambda_0,\lambda_f,\lambda_b,\lambda_h\ge 0$ are the scalar coefficients that balance the relative importance of each category of constraints. The modified (hatted) loss terms are defined as
\begin{equation}
\begin{gathered}
    \hat{\mathcal{L}}_0 := \tfrac{1}{N_0}\,\left\|\,m_0\bigl(u_\theta(\boldsymbol{x}_0)-u_0\bigr)\,\right\|_2^2,
    \\
    \hat{\mathcal{L}}_f := \tfrac{1}{N_f}\,\left\|\,m_f f\bigl(\boldsymbol{x}_f,\mathcal{D}u_\theta(\boldsymbol{x}_f)\bigr)\,\right\|_2^2,
    \\
    \hat{\mathcal{L}}_b := \tfrac{1}{N_b}\,\left\|\,m_b b\bigl(\boldsymbol{x}_b,\mathcal{D}u_\theta(\boldsymbol{x}_b)\bigr)\,\right\|_2^2,
    \\
    \hat{\mathcal{L}}_{h} := \tfrac{1}{N_h}\,\left\|\,m_h\,[\,h\bigl(\boldsymbol{x}_h,\mathcal{D}_h u_\theta(\boldsymbol{x}_h)\bigr) ]_+ \,\right\|_2^2.
    \label{eq:categorised_losses}
\end{gathered}
\end{equation}

Factors $m_0,m_f,m_b,m_h$ are the weights of diagonal matrices applied to each residual component before computing the Euclidean norm. Their role is to provide appropriate scaling to residuals corresponding to different physical quantities, thereby preventing imbalance due to differing magnitudes, and to allow selective emphasis on certain state variables or differential operators in multi-output neural networks. In contrast, the coefficients $\lambda$ in~\eqref{eq:dc-pinns-loss} operate at the categorical level, balancing supervised data, PDE residuals, boundary conditions, and inequality constraints as a whole.  

Often, we face several conditions for inequality constraints simultaneously. In that case, we introduce additional terms of the form $\lambda_h \hat{\mathcal{L}}_h$ for each constraint, leading to a collection of penalties corresponding to multiple inequalities. The set of all such $K_h$ conditions can be denoted as
\begin{equation}
    \mathcal{H}u_\theta := \{\,h_k\bigl(\boldsymbol{x}_{h_k},\mathcal{D}_{h_k} u_\theta(\boldsymbol{x}_{h_k})\bigr)\}_{k=1}^{K_h},\qquad \boldsymbol{x}\in\Omega,
\end{equation}
which may involve multiple, possibly derivative-based, inequality constraints.  

DC-PINNs generalise standard PINNs by simultaneously enforcing PDE dynamics, boundary or initial conditions, and derivative inequality constraints. The framework allows solutions that are consistent with both the governing PDE system and additional derivative-based structural conditions that reflect physical or financial principles, but also introduces additional difficulty as it effectively becomes a multi-objective optimisation problem where the different categories of losses must be balanced to reach a feasible and accurate solution.

\subsection{Multi-Objective Optimisation}
\label{subsec:multi_objective_optimisation}
Selecting weights and constraints-sensitive losses is delicate because large penalties improve constraint satisfaction but can make optimisation ill-conditioned \cite{lu2021physics}. To improve robustness, we employ two automated balancing schemes inspired by \cite{mcclenny2023self, wang2023expert}. The first amplifies the gradients per category to strengthen the derivative-based inequality constraints. The second mitigates multiscale imbalance across the categorised losses in \eqref{eq:categorised_losses}, stabilising the magnitudes of the gradient over epochs given the sparsity induced by \eqref{eq:inequality_loss_hinge}.

\paragraph{Individual Loss Balancing.}
Following a self-adaptive weighting strategy, we attach trainable multiplicative weights $m$ to each categorised loss and update them via gradient ascent in tandem with the network parameters $\theta$:
\begin{equation}
    m_\chi^{(j)}(k{+}1)=m_\chi^{(j)}(k)+\eta_m\,\nabla_{m_\chi^{(j)}}\hat{\mathcal{L}}_\chi(k), \quad \chi\in\{0,b,f,h\}.
\end{equation}
This update increases the influence of losses with larger violations.

\paragraph{Categorised Loss Balancing.}
In parallel, we update the weights per category $\lambda$ using the mean absolute gradient magnitude with respect to $\theta$:
\begin{equation}
\lambda_\chi(k{+}1)=
\begin{cases}
1, & \hspace{-30pt}\text{if }\overline{\lvert \nabla_\theta \hat{\mathcal{L}}_\chi(k)\rvert}=0,\\[3pt]
\lambda_\chi(k)+\nicefrac{\sum_{\chi'} \overline{\lvert \nabla_\theta \hat{\mathcal{L}}_{\chi'}(k)\rvert}}{\overline{\lvert \nabla_\theta \hat{\mathcal{L}}_\chi(k)\rvert}}, & \text{otherwise,}
\end{cases}
\end{equation}
where $\overline{\lvert\cdot\rvert}$ denotes the mean absolute value in the element. We use absolute values, rather than squared ones, to retain sensitivity to rare but severe inequality violations, since most of the elements of $\nabla_\theta \mathcal{L}_h$ are typically zero.

Both mechanisms adjust weights at user-specified intervals based on relative gradient scales, promoting balanced contributions of all loss terms, including unstable inequality losses, during training \cite{mcclenny2023self, wang2023expert}. The sensitivity to $\eta_m$ and $p_m$ is reported in Appendix~\ref{app:sensitivity_analysis_of_adaptive_parameters}

\subsection{Hard-Constraint Formulations for Derivative-Constrained PDEs}
\label{sec:hc_pinns}
We next consider an alternative strategy to DC-PINNs for handling additional constraints, namely a hard-constraint approach. Several practically relevant realisations are based on the augmented Lagrangian methodology \citep{son2023enhanced, lu2021physics} and the classical constrained optimisation theory \citep{nocedal2006numerical, bertsekas2014constrained}.
We treat these approaches as fair baselines because they represent the three most standard families of constrained optimisation in the PINN literature: (i) exact (architectural) enforcement when a closed-form transform exists, (ii) penalty continuation, and (iii) (augmented) Lagrangian updates with dual variables. All target the same feasibility objective as DC-PINNs but differ in how they manage optimisation stiffness and weight selection, which complements the multi-objective discussion in Sec.~\ref{subsec:multi_objective_optimisation}.

\paragraph{Architectural approach (hPINN).}
Pointwise box constraints for boundary conditions ($b$) can be enforced exactly by output transforms in hPINN formulations \cite{lu2021physics},
\begin{equation}
\psi^{(hard)}_\theta=u_{\min}+(u_{\max}-u_{\min})\psi(\boldsymbol{x}).
\end{equation}
This approach is effective for variables whose admissible range is known \emph{a priori}. However, it is generally impractical for derivative-based inequalities such as $u_x\le U$. We include this baseline to represent the strongest form of hard enforcement available when admissible transforms exist, and to make explicit its limited applicability to derivative inequalities.

\paragraph{Penalty method (pen.).}
We penalise violations of derivative inequalities via $[h_\theta]_+$ and employ an outer loop that monotonically increases the weight $\lambda>0$ by scaling it by a factor $\beta>1$ at outer iteration $i$:
\begin{equation}\label{eq:ineq_penalty}
\begin{gathered}
\mathcal{L}^{(\mathrm{pen.})} = \mathcal{L}_0 + \lambda^{(i)}\mathcal{L}_f + \lambda^{(i)}\mathcal{L}_h, \\ \lambda^{(i+1)}=\beta\lambda^{(i)}\ \ (\beta>1).
\end{gathered}
\end{equation}
This is the hard-constraint analogue of a soft hinge penalty, hardened by outer iterations. This baseline is widely used because it is simple to implement, but it can introduce stiff optimisation dynamics as $\lambda^{(i)}$ grows, making comparisons with adaptive weighting particularly informative.

\paragraph{Augmented Lagrangian method (AL).}
Derivative inequalities can also be handled via an augmented Lagrangian loop. The augmented objective with multipliers $\mu_f\in\mathbb{R}^{N_f}$, $\mu_h\in\mathbb{R}^{N_h}$ and $\mu_h\ge 0$ is
\begin{equation}\label{eq:ineq_AL_h}
\begin{gathered}
\mathcal{L}^{(\mathrm{AL})}
= \mathcal{L}_0
+ \mu_f^{\top} f_\theta + \lambda^{(i)}\mathcal{L}_f
+ \mu_h^{\top}[h_\theta]_+ + \lambda^{(i)}\mathcal{L}_h,
\\
\lambda^{(i+1)}=\beta\,\lambda^{(i)}\ (\beta>1).
\end{gathered}
\end{equation}
with projected dual-ascent updates and a multiplicatively increased penalty parameter,
\begin{equation}\label{eq:ineq_AL_update_h}
\begin{gathered}
\mu_f^{(i+1)}=\mu_f^{(i)}+\lambda^{(i)} f_\theta^{(i+1)},\\
\mu_h^{(i+1)}=\mu_h^{(i)}+\lambda^{(i)}[h_\theta^{(i+1)}]_+.
\end{gathered}
\end{equation}
Thus, hPINNs can extend by augmenting the treatment of inequalities while retaining hard architectural enforcement (hPINN (AL)). In addition, all constraints can be unified into a single AL loop (AL-PINNs). We use AL-based baselines because they are a principled alternative to fixed penalty weights and are commonly adopted in constrained PINNs; however, their outer-loop updates can be sensitive to stopping criteria and can prioritise feasibility at the expense of data-fit under a fixed compute budget.

To our knowledge, no prior PINN study has focused on a fully unified hard-constraints formulation that simultaneously addresses multiple equality and inequality constraints. Beyond prior work on equality constraints \cite{lu2021physics,son2023enhanced}, our formulation integrates multiple inequalities, tightening feasibility coupling, but also increasing the ill-conditioning of the optimisation landscape and, consequently, the risk of convergence difficulties.

\section{Experimental Design}
\label{section: Experimental Design}
\subsection{Neural Network Setting and Training Configuration}
\label{Neural Network Setting and Training Configuration}
In the experiments, the network architecture $\psi$ is a fully connected feedforward neural network with hyperbolic tangent activation functions to ensure smooth differentiability. The network parameters are initialised using the Glorot scheme \cite{glorot2010understanding}, and the optimisation is performed using Adam \citep{kingma2014adam} with regularisation of the weight decay, following previous work \citep{raissi2019physics, wang2023expert}. Training data consist of uniformly sampled points for initial and boundary conditions, while square mesh grids are used for PDE residuals and derivative inequality constraints. The accuracy of the model is evaluated by computing the errors between predictions and reference solutions on dense mesh grids. Detailed analytical formulations, network architectures, and training settings for each problem are provided in Appendix~\ref{app:heat_impl}, Appendix~\ref{app:lv_impl:training}, and Appendix~\ref{app:ns_impl:training}.

\subsection{Baselines and fairness}
Standard PINNs~\cite{raissi2019physics} do not encode derivative-based inequality constraints. We therefore add \emph{PINNs-Ineq.}, which augments PINNs with a hinge penalty $\mathcal{L}_h$ while fixing additional weights ($\lambda_\chi\!\equiv\!1$, $m_\chi\!\equiv\! I$). \emph{DC-PINNs} generalise this learning per component coefficients $\lambda_\chi$ and per sample scale $m_\chi$ (Section~\ref{sec:dc-pinns}). We also report ablations: \emph{no $\lambda$} ($\lambda_\chi=1$), \emph{no $m$} ($m_\chi=I$), and \emph{static} (both initialized then frozen). For hard-constraint baselines (Section~\ref{sec:hc_pinns}): \emph{hPINN (pen.)} embeds boundary conditions (hard-$b$) and enforces inequalities through a penalty homotopy; \emph{hPINN (AL)} again uses hard-$b$ with an augmented Lagrangian for inequalities; \emph{AL-PINNs} applies an augmented Lagrangian to both equalities and inequalities without hard-$b$. All baselines follow the effective hyperparameters of the original paper~\cite{lu2021physics}. The explored ranges and final settings appear in the Appendix~\ref{app:sensitivity_analysis_of_adaptive_parameters}.

\paragraph{Metrics.}
Accuracy is measured by root mean square error (RMSE). For each component $\chi\in\{0,b,f,h\}$ on the dense validation grid $\{\boldsymbol{x}_i\}_{i=1}^{N_\chi^{\mathrm{val}}}$ with task-specific error $e_\chi(\boldsymbol{x}_i)$ (data, boundary, PDE residual, or derivative error), we report
\begin{equation}
    \mathrm{E}_\chi=\mathrm{RMSE}_\chi=\left(\tfrac{1}{N_\chi^{\mathrm{val}}}\Sigma_{i=1}^{N_\chi^{\mathrm{val}}} e_\chi(\boldsymbol{x}_i)^2\right)^{1/2}.
\end{equation}

\paragraph{Environment.} All models involving differentiable operators were implemented using JAX and Flax \citep{jax2018github, flax2020github}, which enable efficient and exact computation of derivatives via automatic differentiation. Training was carried out in Google Colab \cite{googlecolab} using TPUs (v6e-1) with XLA compilation and 173 GB of high-bandwidth memory. On average, each training run required approximately 5 (Heat) to 100 (Navie-Stokes) seconds for runtime and 2-3 minutes for compiling on a TPU v6e-1 instance, with peak memory usage close to 120 GB during the largest experiments.

\section{Numerical Experiments}
\label{sec:numerical_experiments}
To demonstrate the effectiveness of the proposed framework, we performed numerical experiments on benchmark problems in heat diffusion, volatility-surface calibration in quantitative finance, and incompressible flow with complex geometries and non-linear constraints. We compare our approach with existing PINN-based methods and include derivative profile comparisons of trained networks.

\subsection{One-dimensional Heat Equation in Thermodynamics}
The heat equation is a classic parabolic PDE \cite{cannon1984one} and a standard testbed to evaluate the robustness of PINNs. Consider an infinitesimally thin steel beam heated at its centre. The heat in the centre spreads along the beam while the ends are held at zero temperature, so the temperature decays to zero as $t\to\infty$. The problem in $\boldsymbol{x}=(x,t)$ is
\begin{equation}
f(x,t)=\pdv{u}{t}-\lambda\,\pdv[2]{u}{x},
\label{eq:pde_1d_heat_equation}
\end{equation}
\begin{equation}
\begin{gathered}
\text{s.t.}\quad u(x,0)=\sin(\pi x),\;\;u(0,t)=u(1,t)=0,\\
\qquad\qquad\pdv[2]{u}{x}\le 0,\quad \pdv{u}{t}\le 0,
\label{eq:constraints_1d_heat_equation}
\end{gathered}
\end{equation}
for $x,t\in[0,1]$. The analytical solution is $u(x,t)=e^{-\lambda\pi^2 t}\sin(\pi x)$. The derivative constraints in \eqref{eq:constraints_1d_heat_equation} reflect that the second spatial derivative and the time derivative are non-positive throughout the domain: the initially peaked profile remains concave in space while decaying in time, and set $\lambda=0.1$.

\begin{figure}[htbp]
  \includegraphics[width=\columnwidth]{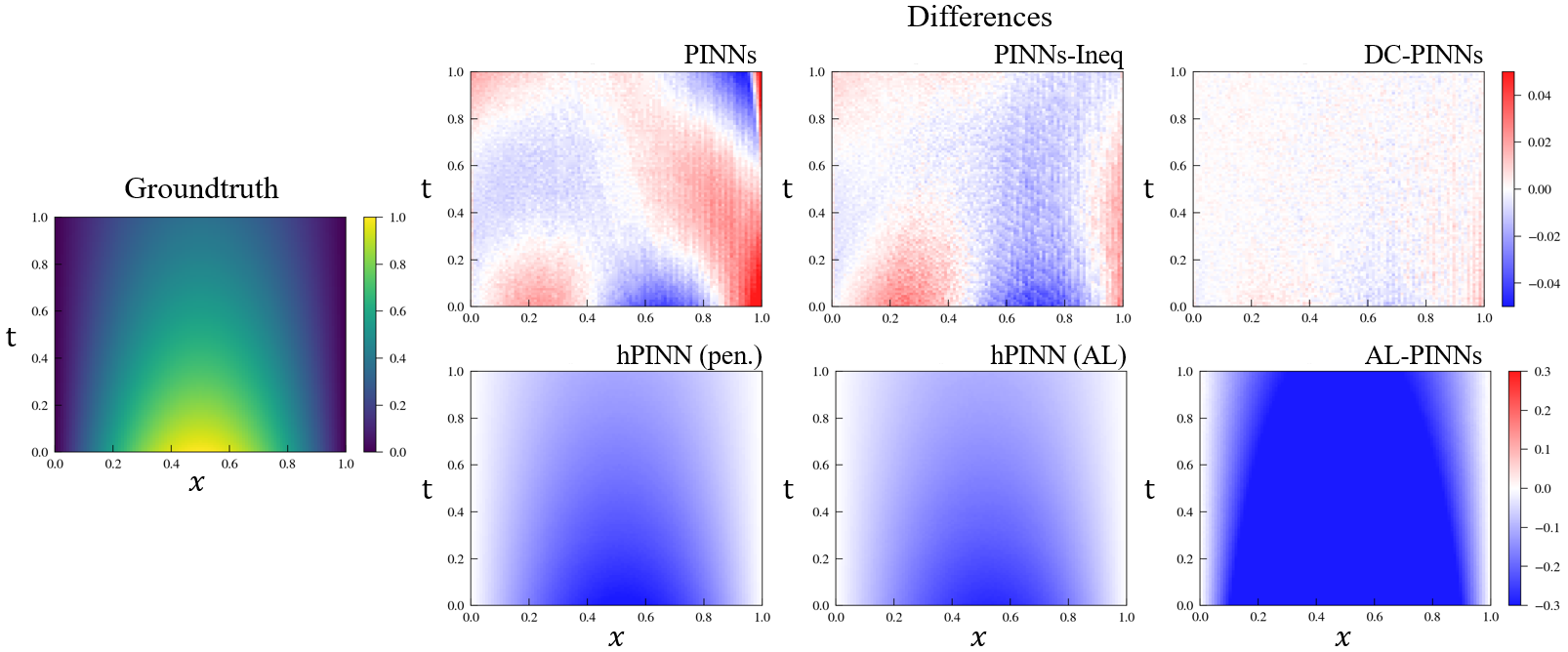}
  \caption{Predicted solutions for the one-dimensional heat equation. Left: ground-truth magnitude. Right: model–ground-truth difference with a diverging scale.}
  \label{fig:DC-PINNs_1DHeat}
\end{figure}
Fig.~\ref{fig:DC-PINNs_1DHeat} shows that DC-PINNs reproduce the smooth diffusive profile of the analytical solution with minimal artefacts. Standard PINNs exhibit noisy oscillations and boundary bias, and adding fixed inequality terms only partly suppresses them. hPINN and AL-PINNs appear over-smoothed, losing temporal variation through excessive constraint enforcement. In contrast, DC-PINNs maintain both spatial smoothness and temporal decay, yielding a near-zero residual map that confirms faithful satisfaction of $\partial_t u=\partial_{xx}u$ across the domain.

\begin{table*}[tp]
\centering
\caption{Validation (RMSE: \textbf{bold} is best) for the heat equation.}
\label{tab:heat_models}
\begin{tabular}{lccc ccccc}
\toprule
& \multicolumn{3}{c}{Conditions} & \multicolumn{5}{c}{Accuracy} \\
Model & Ineq. & Hard-$b$ & AL & {$\mathrm{E}_u$} & {$\mathrm{E}_f$} & {$\mathrm{E}_b$} & {$\mathrm{E}_{h_{xx}}$} & {$\mathrm{E}_{h_t}$} \\
\midrule
PINNs & -- & -- & -- & 0.013 & 0.032 & 0.027 & 0.677 & 0.068 \\
PINNs+Ineq. & \checkmark & -- & -- & 0.010 & 0.035 & 0.010 & 0.522 & 0.037 \\
hPINN (pen.) & \checkmark & \checkmark & -- & 0.144 & 0.010 & \textbf{0.000} & 1.433 & 0.142 \\
hPINN (AL) & \checkmark & \checkmark & \checkmark & 0.131 & 0.010 & \textbf{0.000} & 1.310 & 0.129 \\
AL-PINNs & \checkmark & -- & \checkmark & 0.463 & 0.008 & 0.004 & 4.572 & 0.454 \\
DC-PINNs & \checkmark & -- & -- & \textbf{0.002} & \textbf{0.006} & 0.003 & \textbf{0.091} & \textbf{0.009} \\
\bottomrule
\end{tabular}
\end{table*}

\begin{figure}[htbp]
  \includegraphics[width=\columnwidth]{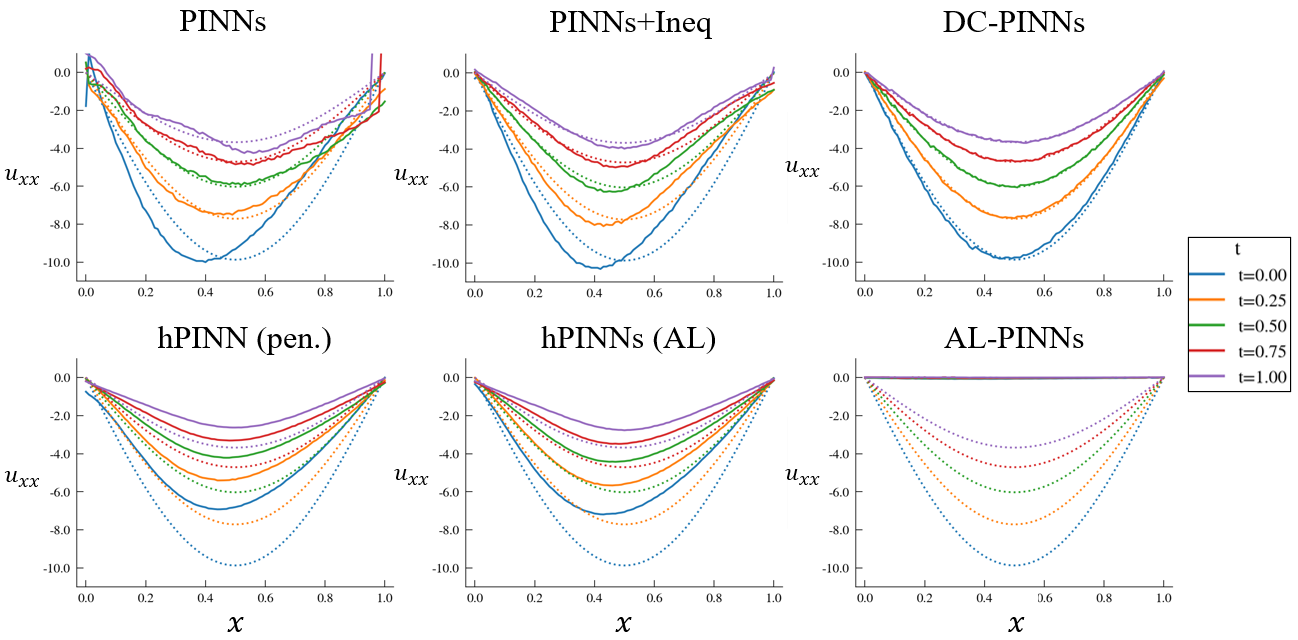}
  \caption{Second-derivative profiles with respect to $x$ at multiple time snapshots. Dashed: analytical solution.}
  \label{fig:DC-PINNs_1DHeat_sens_xx}
\end{figure}

Table~\ref{tab:heat_models} indicates that DC-PINNs achieve the best RMSE under the initial condition, the residual interior PDE, and the boundary (except for the hard-boundary treatments), while simultaneously attaining lower derivative RMSEs ($E_{h_{xx}}$, $E_{h_t}$). By contrast, AL-based variants drive inequality-violation metrics to zero but exhibit larger derivative errors, showing that exact feasibility alone does not guarantee accurate curvature or temporal dynamics. The qualitative second-derivative profiles in Fig.~\ref{fig:DC-PINNs_1DHeat_sens_xx} corroborate this: DC-PINNs track the analytic curvature closely across space–time without spurious boundary oscillations. In general, DC-PINNs balance feasibility and accuracy, whereas purely hard treatments tend to over-constrain the solution away from the data and PDE signals.

To further examine the influence of inequality constraints in DC-PINNs, Fig.~\ref{fig:DC-PINNs_1DHeat_sens_xx} compares the learnt second derivative profiles with the analytical curves over time. Standard PINNs deviate markedly near the boundaries, including sign reversals that violate $u_{xx}\le 0$. Fixed-inequality PINNs stabilise the profiles but retain magnitude bias, indicating that static penalties cannot guarantee global compliance. AL- and hPINNs produce smoother, yet overly flattened curvature, suppressing temporal evolution. AL-PINNs almost collapse to zero curvature. In contrast, DC-PINNs preserve the parabolic shape and maintain non-positivity of $u_{xx}$ everywhere, confirming that embedding derivative inequalities in the loss is essential for both physical realism and temporal coherence.

\subsection{Quantitative Analysis and Ablation}
\label{subsec:quant}
We analyse the impact of the enforcement of inequality and adaptive balancing through controlled ablations. The hyperparameter sensitivity to $\eta_m\in\{10^{-4},10^{-3},10^{-2}\}$ and $p_m,p_\lambda\in\{10,100,1000\}$ is summarised in Tables~\ref{tab:sensitivity_all_heat}–\ref{tab:sensitivity_all_ns} in Appendix~\ref{app:sensitivity_analysis_of_adaptive_parameters}. The results consistently favour $\eta_m\approx10^{-3}$-$10^{-2}$ and moderate update intervals, indicating robustness to tuning.

\begin{table*}[tp]
\centering
\caption{Ablation on validation (RMSE: \textbf{bold} is best) and adaptive balancing in the heat equation. $\lambda$: component-wise; $m$: sample-wise.}
\label{tab:ablation}
\begin{tabular}{l c c c c c c c c}
\toprule
& \multicolumn{3}{c}{Conditions} & \multicolumn{5}{c}{Accuracy} \\
Model & Ineq. & $\lambda$ & $m$ & $E_0$ & $E_f$ & $E_b$ & $E_{h_{xx}}$ & $E_{h_{t}}$ \\
\midrule
PINNs & -- & -- & -- 
& 0.013 & 0.032 & 0.027 & 0.678 & 0.068 \\
PINNs-Ineq (fixed) & \checkmark & -- & -- 
& 0.010 & 0.036 & 0.010 & 0.524 & 0.037 \\
DC-PINNs (no $\lambda$) & \checkmark & -- & \checkmark 
& 0.012 & 0.056 & 0.020 & 0.501 & 0.050 \\
DC-PINNs (no $m$) & \checkmark & \checkmark & -- 
& 0.004 & 0.008 & 0.006 & 0.160 & 0.016 \\
DC-PINNs (full) & \checkmark & \checkmark & \checkmark 
& \textbf{0.002} & \textbf{0.006} & \textbf{0.003} & \textbf{0.091} & \textbf{0.009} \\
\bottomrule
\end{tabular}
\end{table*}
The ablation in Table~\ref{tab:ablation} clarifies the contributions of inequality enforcement ($L_h$) and adaptive balance ($\lambda,m$). Introducing $L_h$ alone improves the satisfaction of the constraint and reduces RMSE relative to the unconstrained baseline, indicating that inequality enforcement directly improves physical fidelity. The greatest gains occur when the balance between components ($\lambda$) and sample-wise ($m$) balance is active, producing up to an order of magnitude reduction in RMSE and constraint errors. Removing either element degrades performance, particularly in $E_{h_t}$, suggesting that temporal derivatives are sensitive to unbalanced weighting. Static weights do not maintain consistent constraint satisfaction, underscoring the value of dynamic adaptation.

\subsection{Volatility Surface Calibration in Finance}
\label{subsec:vol_surface_calib}
The volatility-surface calibration problem is an inverse PDE setting in quantitative finance, where the objective is to recover all the option price data from the observed data \cite{dupire1994pricing}. The model satisfies
\begin{equation}
\begin{gathered}
f(x,t)=\pdv{u}{t}-\tfrac{1}{2}\sigma(x,t)\,x^2\pdv[2]{u}{x}+r\,x\,\pdv{u}{x},\\
\sigma(x,t)=\sigma_A+\tfrac{A}{x}+Bt,
\end{gathered}
\end{equation}
\begin{equation}
\begin{gathered}
\text{s.t.}\quad u_{t=0}=(s_0-x)^+,\; u_{x=0}=s_t,\\
\qquad\pdv{u}{x}\le 0,\;\pdv[2]{u}{x}\ge 0,\;\pdv{u}{t}\ge 0,
\end{gathered}
\end{equation}
where $u$ is the European call option price. Synthetic data are generated from the analytical $\sigma(\cdot)$ with $\sigma_A=0.2$, $A=0.1$, $B=0.2$ \citep{kim2021reconstruction,boyle2000volatility}. The domain is $x\in[0,2]$, $t\in[0,1]$ with $s_0=1.0$, $r=0.05$, and Gaussian noise added to Monte Carlo prices (Appendix~\ref{app:lv_method_impl}). During training, $u:=\psi_\theta(x,t)$. The no-arbitrage inequalities $\partial_x u\le 0$, $\partial_{xx}u\ge 0$, and $\partial_t u\ge 0$ ensure economically valid surfaces. Embedding these constraints in the objective enables DC-PINNs to produce PDE-consistent, arbitrage-free surfaces.

\begin{figure}[htbp]
  \includegraphics[width=\columnwidth]{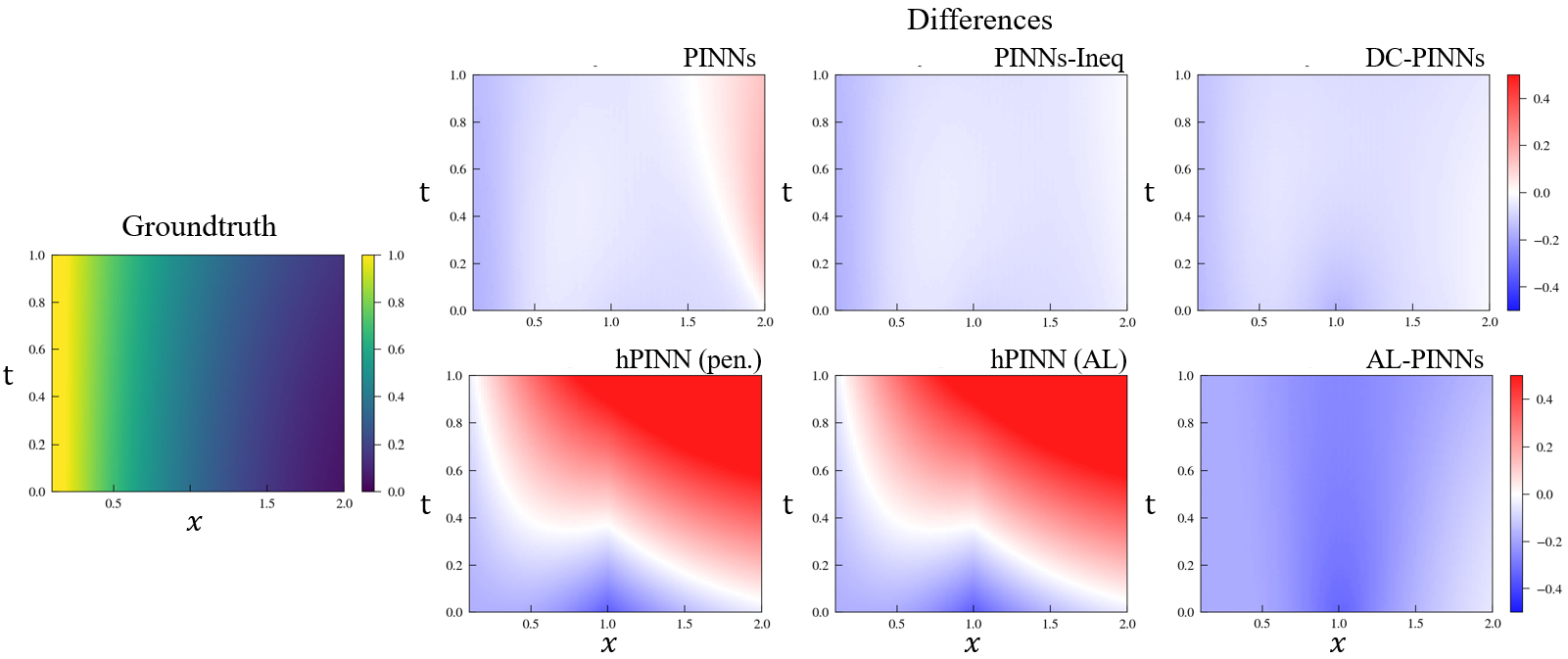}
  \caption{Predicted solutions for the volatility-surface calibration. Left: ground-truth magnitude. Right: model–ground-truth difference with a diverging scale.}
  \label{fig:DC-PINNs_LV}
\end{figure}
FIG.~\ref{fig:DC-PINNs_LV} shows reconstructed surfaces. DC-PINNs yield smooth, monotonic profiles consistent with the ground truth, whereas unconstrained and fixed-penalty PINNs exhibit locally irregular curvature near strike extremes.

\begin{table*}[tp]
\centering
\caption{Validation (RMSE: \textbf{bold} is best) for the volatility surface calibration.}
\label{tab:lv_results}
\begin{tabular}{l c c c c c c c c c}
\toprule
& \multicolumn{3}{c}{Conditions} & \multicolumn{6}{c}{Accuracy} \\
Model & Ineq. & Hard-$b$ & AL & {$E_0$} & {$E_b$} & {$E_f$} & {$E_{h_x}$} & {$E_{h_{xx}}$} & {$E_{h_t}$} \\
\midrule
PINNs                & --         & --         & --         & 0.002 & \textbf{0.002} & \textbf{0.002} & 0.154 & 0.065 & \textbf{0.000} \\
PINNs+Ineq. (fixed)  & \checkmark & --         & --         & 0.002 & 0.003 & 0.004 & 0.098 & 0.019 & 0.029 \\
hPINN (pen.)         & \checkmark & \checkmark & --         & 0.477 & 0.007 & 0.878 & 0.533 & 0.423 & \textbf{0.000} \\
hPINN (AL)           & \checkmark & \checkmark & \checkmark & 0.477 & 0.007 & 0.878 & 0.533 & 0.419 & \textbf{0.000} \\
AL-PINNs             & \checkmark & --         & \checkmark & 0.002 & 0.003 & 0.002 & 0.057 & \textbf{0.005} & 0.032 \\
DC-PINNs             & \checkmark & --         & --         & \textbf{0.001} & \textbf{0.002} & \textbf{0.002} & \textbf{0.053} & 0.095 & \textbf{0.000} \\
\bottomrule
\end{tabular}
\end{table*}
Table~\ref{tab:lv_results} summarises the calibration performance. DC-PINNs achieve the lowest overall RMSE, particularly in the initial and boundary conditions, while maintaining RMSEs of zero or near zero constraints for $\partial_x u$, $\partial_{xx}u$, and $\partial_t u$. Fixed-penalty PINNs moderately improve over the unconstrained baseline, yet still violate constraints near maturity boundaries. AL- and hPINNs enforce hard constraints, but show optimisation instability and poorer fits. Adaptive balancing in DC-PINNs ensures stable convergence and joint enforcement of derivative constraints, improving both data fidelity and physical admissibility.

\begin{figure}[htbp]
  \includegraphics[width=\columnwidth]{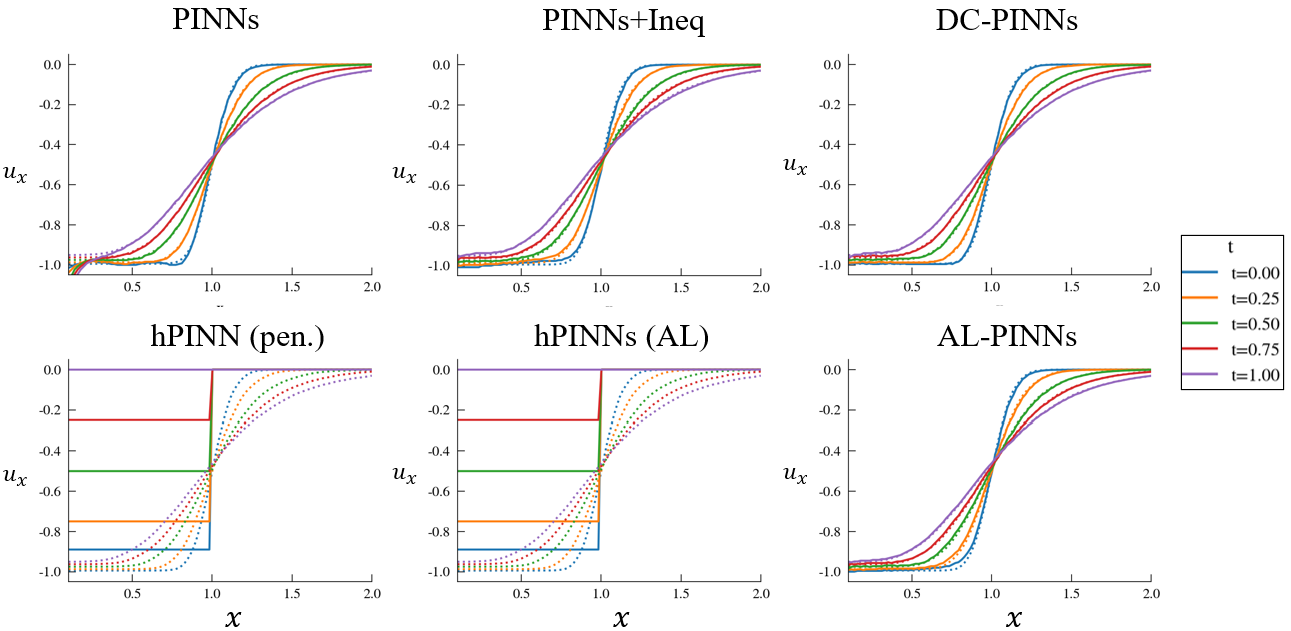}
  \caption{Spatial derivative $\partial u/\partial x$ at multiple time snapshots. Dashed: numerical reference.}
  \label{fig:DC-PINNs_LV_grads_x}
\end{figure}
FIG.~\ref{fig:DC-PINNs_LV_grads_x} shows $\partial u/\partial x$ over time. Standard PINNs and fixed-inequality variants capture the overall trend but deviate near the strike boundaries. DC-PINNs remain closer to the reference in the outer regions, consistent with the lower $E_{h_x}$ in Table~\ref{tab:lv_results}. AL- and hPINNs introduce discontinuities around $x\approx 1$, producing piecewise or over-smoothed gradients that break the expected monotonicity. AL-PINNs recover the general trend but underestimate slopes in the interior, indicating excessive penalisation of higher-order terms.

\begin{figure}[htbp]
  \includegraphics[width=\columnwidth]{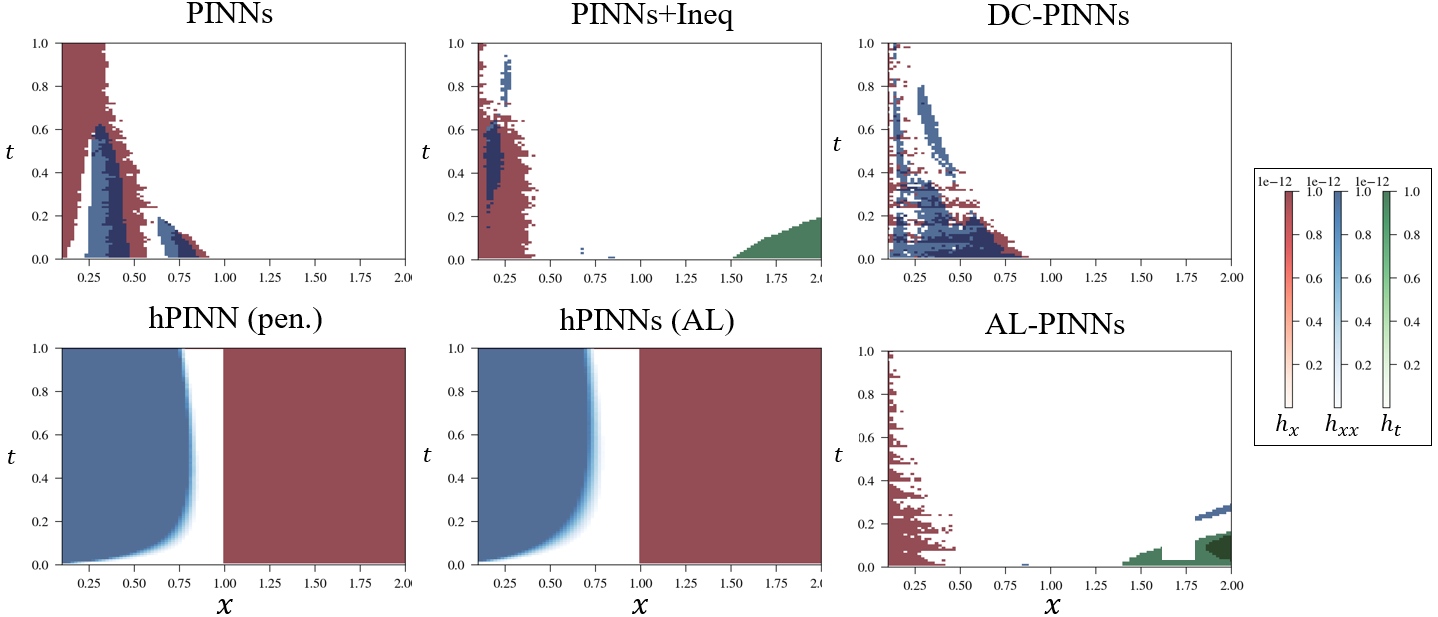}
  \caption{Inequality-violation maps for learned local volatilities; dashed lines indicate exact references.}
  \label{fig:DC-PINNs_LV_arbs}
\end{figure}

As in the other experiments, derivative profile analyses (Fig.~\ref{fig:DC-PINNs_LV_arbs}) show that DC-PINNs capture the solution while satisfying nonlinear constraints, outperforming traditional PINNs. The framework is versatile and applies to data-driven calibration of PDE parameters.

\subsection{Navier–Stokes Equations in Fluid Dynamics}
We now consider the incompressible Navier–Stokes equations, a challenging benchmark in fluid dynamics \cite{raissi2019physics}. The flow past a 2D circular cylinder (von Kármán vortex street) is modelled. The governing equations in convective form are
\begin{equation}
\begin{cases} 
f^u = \pdv{u}{t} + \mu_1\left(u \pdv{u}{x} + v \pdv{u}{y}\right) + \pdv{p}{x} - \mu_2\left(\pdv[2]{u}{x} + \pdv[2]{u}{y}\right) = 0\\
f^v = \pdv{v}{t} + \mu_1\left(u \pdv{v}{x} + v \pdv{v}{y}\right) + \pdv{p}{y} - \mu_2\left(\pdv[2]{v}{x} + \pdv[2]{v}{y}\right) = 0
\end{cases}
\label{eq: pde Navier Stokes}
\end{equation}

\begin{equation}
\begin{gathered}
\mathrm{s.t.} \quad \left.\left(u,v\right)\right\rvert_{x=-15} = \left.\left(u,v\right)\right\rvert_{y=-8,8} = \left(1,0\right), \\ \left.\pdv{u}{x},\pdv{v}{x}\right\rvert_{x=25} = \left.p\right\rvert_{x=-15} = 0,\\
u = v = 0 \text{ on the cylinder surface},
\label{eq: boundaries Navier Stokes}
\end{gathered}
\end{equation}
\begin{equation}
\begin{gathered}
\nabla \cdot \mathbf{V} = \pdv{u}{x} + \pdv{v}{y} = 0, \;\;\;\;\; 
|\nabla p| \leq \tfrac{1}{2}\rho U^2 \quad \label{eq: constraints Navier Stokes}
\end{gathered}
\end{equation}
where $\mathbf{V}=(u,v)$, $p$ is pressure, and $\mu_1=1$, $\mu_2=1/\mathrm{Re}$ with $\mathrm{Re}$ the Reynolds number. We treat this as PDE discovery, estimating $\mu=(\mu_1,\mu_2)$ from the data. The flow past a cylinder of diameter $D=1$ at $\mathrm{Re}=100$ is simulated. Training samples of $\mathbf{V}$ are drawn at 5{,}000 random points in $x\in[1,8]$, $y\in[-2,2]$ (cylinder centred at the origin), and $t\in[0,20]$ in periodic steady flow. Incompressibility ($\nabla\cdot\mathbf{V}=0$) enforces mass conservation \cite{peyret2012computational}; the bound of the pressure gradient follows the Bernoulli scale with $\rho=U=1$ \cite{batchelor2000introduction}. Reference solutions are computed with the Nektar++ spectral / hp element code \citep{moxey2020nektar++} (Appendix~\ref{app:ns_method_impl}).

\begin{table*}[tp]
\centering
\caption{Validation (RMSE: \textbf{bold} is best) for Navier–Stokes equations.}
\label{tab:ns_results}
\begin{tabular}{l cc ccccc cc}
\toprule
& \multicolumn{2}{c}{Conditions} & \multicolumn{5}{c}{Accuracy} & \multicolumn{2}{c}{$\mu$} \\
Model & Ineq. & AL & {$E_{0_u}$} & {$E_{0_v}$} & {$E_{f_u}$} & {$E_{f_v}$} & {$E_{h_{\nabla p}}$} & {$\mu_1$} & {$\mu_2$} \\
\midrule
PINNs                 & --         & --         & 0.395 & 0.395 & 0.018 & 0.019 & 0.159 & \textbf{0.977} & \textbf{0.010} \\
PINNs+Ineq. (fixed)   & \checkmark & --         & 0.395 & 0.391 & 0.020 & 0.019 & 0.154 & \textbf{0.977} & 0.011 \\
hPINN (pen.)          & \checkmark & --         & 0.350 & \textbf{0.304} & \textbf{0.004} & \textbf{0.004} & 0.165 & 0.758 & 0.020 \\
AL-PINNs            & \checkmark & \checkmark & \textbf{0.343} & 0.350 & \textbf{0.004} & 0.005 & \textbf{0.142} & 0.875 & 0.012 \\
DC-PINNs              & \checkmark & --         & 0.395 & 0.391 & 0.018 & 0.016 & 0.153 & \textbf{0.977} & \textbf{0.010} \\
\bottomrule
\end{tabular}
\end{table*}

For the two-dimensional cylinder wake, Table~\ref{tab:ns_results} shows that the relative ordering depends on which objective is prioritised. AL-PINNs attain the lowest pressure-constraint error $E_{h_{\nabla p}}$ and the lowest $E_{0_u}$, while the hPINN (pen.) baseline yields the lowest $E_{0_v}$ and the smallest PDE residuals $(E_{f_u},E_{f_v})$. DC-PINNs reduce $E_{h_{\nabla p}}$ relative to unconstrained PINNs and recover $\mu_2$ close to its reference value, but they do not improve $E_0$ or $E_f$ over the best-performing baselines in this benchmark. We therefore interpret the Navier--Stokes results primarily as evidence of improved constraint handling, with an explicit accuracy--feasibility trade-off under a fixed training budget. A systematic study of the dependence on network capacity and training time is left for future work. Across hyperparameter sweeps (Table~\ref{tab:sensitivity_all_ns}), the recovered viscosity consistently converges to $\mu_2\approx 0.01$ for $\nu=10^{-2}$ ($\mathrm{Re}=100$), confirming accurate inference of the underlying physics (Appendix~\ref{app:sensitivity_analysis_of_adaptive_parameters}).

\begin{figure}[htbp]
  \includegraphics[width=\columnwidth]{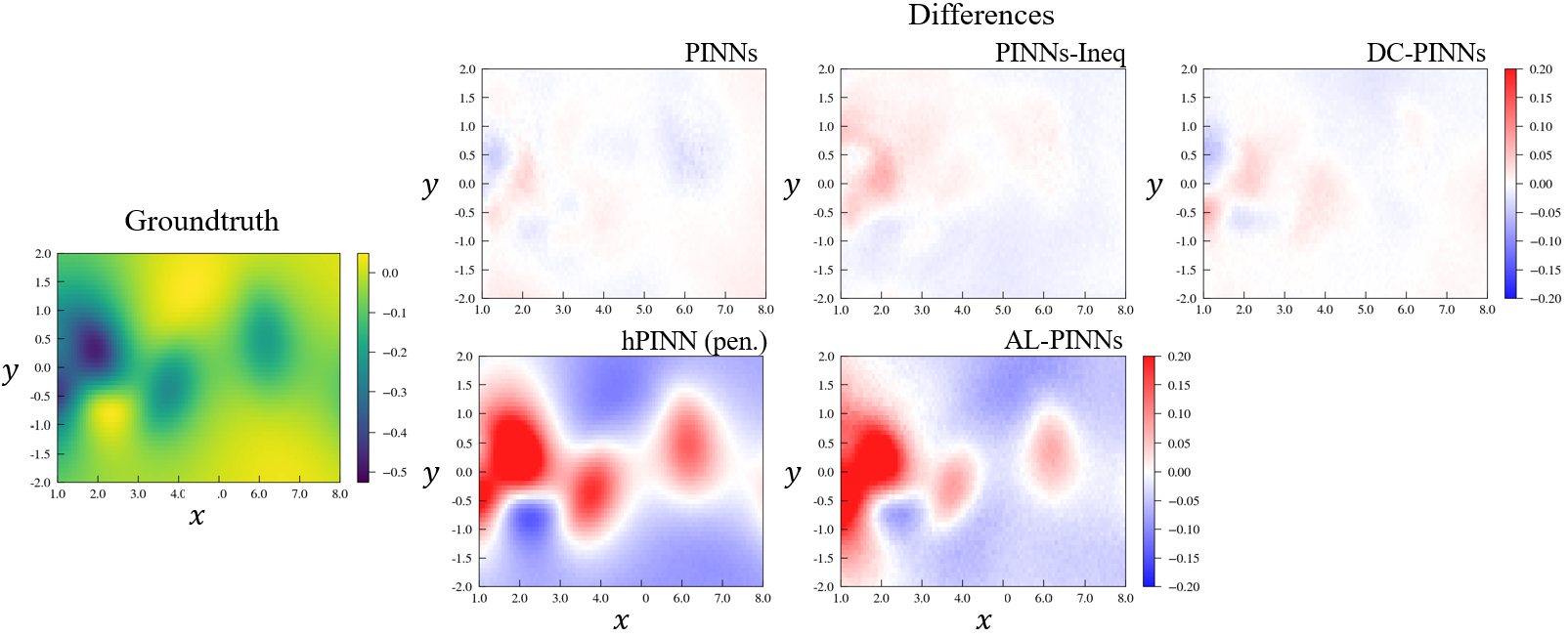}
  \caption{Predicted pressure $p(x,y,t)$ (mean-adjusted) at $t=10$ for the Navier–Stokes case. Left: ground-truth magnitude. Right: model–ground-truth difference with a diverging scale.}
  \label{fig:DC-PINNs_NS_ps}
\end{figure}
FIG.~\ref{fig:DC-PINNs_NS_ps} compares reconstructed pressure fields in a representative phase of the shedding cycle. Most methods reproduce the gross wake symmetry, and the snapshot-level differences are subtle; consequently, we use the figure as qualitative context and rely on Table~\ref{tab:ns_results} for quantitative assessment. As discussed above, these results reinforce the central trade-off in this benchmark: improved feasibility of the derivative constraint does not necessarily coincide with the lowest data error or residual.

\section{Discussions}
\subsection{Computational complexity}
We report wall-clock training times for all methods on the three benchmark problems using identical multilayer perceptron (MLP) architectures and data pipelines as in the accuracy experiments. For the heat equation, the network consists of two hidden layers with 100 neurones each, trained on grids of $31\times31$ interior points, $1001$ initial points and $2\times1001$ boundary points. The volatility surface problem employs a fully connected network with four hidden layers of 50 neurones each, using an interior grid of $50\times50$, $1000$ observed prices, and $2\times1001$ boundary points. For the two-dimensional Navier-Stokes case, we used eight hidden layers with 20 neurones per layer and $N_\chi = 5000$ collocation samples. Table~\ref{tab:complexity_time} summarises the training times per run, measured in seconds until the reported evaluation, under identical JIT compilation and data loading settings.

\begin{table}[t]
\centering
\small
\caption{Wall-clock training time (s) across benchmark problems.}
\label{tab:complexity_time}
\begin{tabular}{lccc}
\hline
Method & \makecell{Heat\\Problem} & \makecell{Volatlity\\Surface} & \makecell{Navier\\–Stokes} \\
\hline
PINNs & 2.16 & 3.93 & 58.52 \\
PINNs+Ineq & 2.19 & 4.05 & 82.14 \\
hPINNs (pen.) & 2.82 & 3.85 & 42.55 \\
hPINNs (AL) & 2.88 & 4.36 & 52.68 \\
AL-PINNs & 2.02 & 3.51 & --- \\
\textbf{DC-PINNs} & \textbf{4.41} & \textbf{6.17} & \textbf{97.10} \\
\hline
\end{tabular}
\end{table}

Across all experiments, DC-PINNs incur roughly a $1.5$–$2\times$ training-time overhead relative to unconstrained PINNs, but this cost yields systematic gains in physical consistency and, in most cases, accuracy. For the Heat problem, second derivative violations $(u_{xx}\ge0,u_t\ge0)$ are markedly reduced; in the case of the volatlity surface, boundary and derivative errors decline, producing price surfaces consistent with Dupire’s constraints; and in Navier–Stokes, velocity and pressure errors remain comparable while divergence and momentum residuals tighten, improving stability. Overall, the results show that DC-PINNs are most beneficial when derivative constraints dominate the ill-posedness (as in Heat and Volatility), whereas for coupled multi-field systems (Navier–Stokes) they primarily enhance consistency rather than headline accuracy, with AL- or hPINN variants offering faster but occasionally less stable alternatives.

\subsection{Evaluations}
We evaluated four main aspects of performance, namely accuracy, stability, efficiency, and violation. Accuracy refers to the final validation error obtained at the end of training. Stability is measured through the normalised total variation of the validation error trajectory, which captures oscillations during optimisation. Efficiency is assessed by the time required for the error to fall to half of its initial value, together with the area under the improvement curve, both measured with respect to wall clock time. The violation metric quantifies the degree to which physical or structural constraints remain unsatisfied at the end of the training.

Let $\mathbf{e}=[e_1,\dots,e_n]$ denote the validation errors recorded at times $\mathbf{t}=[t_1,\dots,t_n]$ with $e_{\min}=\min_i e_i$, $e_{\max}=\max_i e_i$, and a small positive constant $\varepsilon$. The final accuracy is given by $e_{\text{final}}=e_n$. The stability measure is expressed as
\begin{equation}
\mathrm{TVN}=\frac{\sum_{i=1}^{n-1}\lvert e_{i+1}-e_i\rvert}{e_{\max}-e_{\min}+\varepsilon},
\end{equation}
while efficiency is expressed as
\begin{equation}
\mathrm{nAUC}=\frac{1}{t_n-t_1+\varepsilon}\int_{t_1}^{t_n}\frac{e_{\max}-e(t)}{e_{\max}-e_{\min}+\varepsilon}\,dt.
\end{equation}
This definition rewards models that sustain low error values over time. The violation score is calculated as the average magnitude of the constraint residuals on the validation grid. Improvements are reported as percentage changes relative to PINNs. For metrics to be minimised: $100\,(m_{\mathrm{PINNs}}-m)/\max(m_{\mathrm{PINNs}},\varepsilon)$; for metrics to be maximised: $100\,(m-m_{\mathrm{PINNs}})/\max(m_{\mathrm{PINNs}},\varepsilon)$. The methods are ranked per metric and the overall performance is the Borda sum of ranks.

\begin{figure}[htbp]
\centering
\includegraphics[width=\columnwidth]{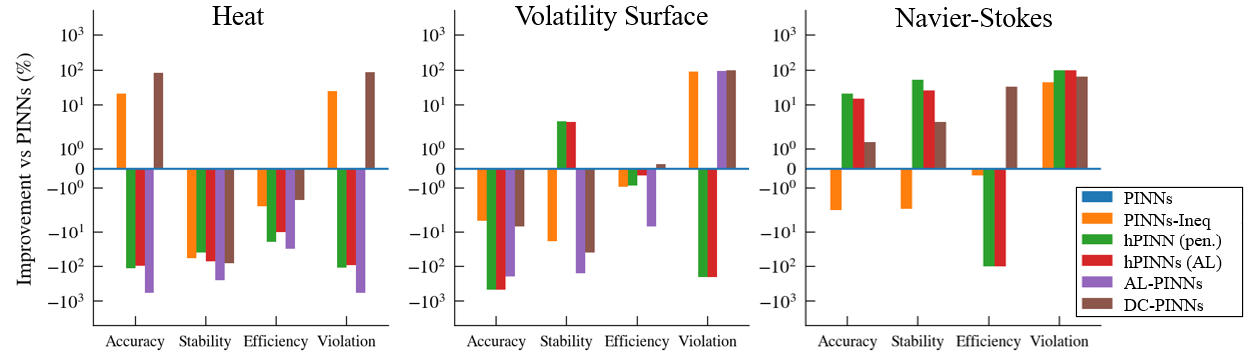}
\caption{Percentage improvement with respect to PINNs on a symmetric logarithmic scale. Positive values indicate improvement relative to the baseline.}
\label{fig:dc_metrics_bar}
\end{figure}
The aggregated results highlight a consistent pattern across tasks: DC-PINNs reduce constraint violations and damp oscillatory training behaviour, but they do so with a measurable cost in computational efficiency. In particular, DC-PINNs typically require additional derivative evaluations and longer wall-clock time, which is reflected in a lower efficiency score even when the final accuracy is comparable. Moreover, accuracy improvements are task-dependent: in Navier--Stokes, for example, AL- and hPINN baselines achieve lower data error and residual metrics. We therefore present DC-PINNs as a method that prioritises feasibility and stability, rather than a uniformly dominating replacement for existing approaches.

\begin{figure}[htbp]
\centering
\includegraphics[width=\columnwidth]{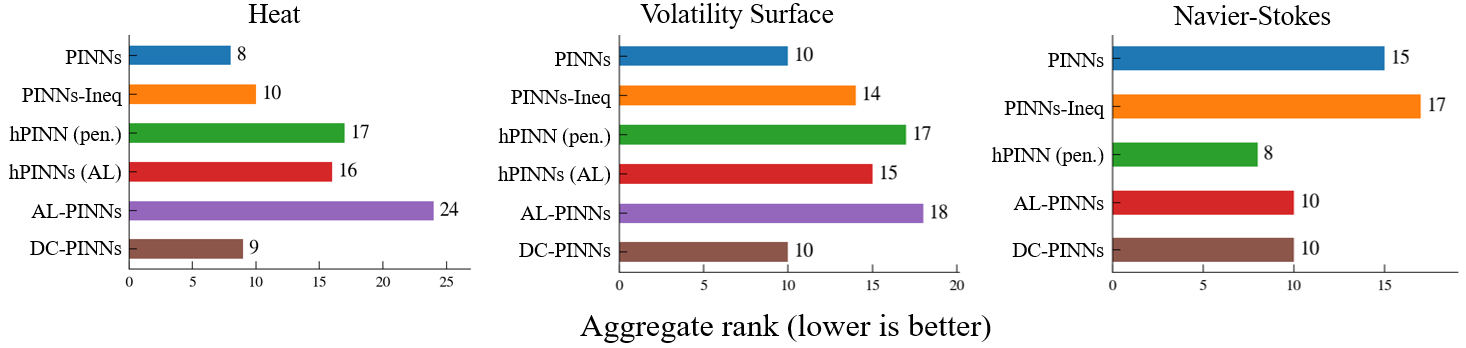}
\caption{Aggregate ranks across all metrics. Left Heat-PDE, centre Volatility Surface PDE, right Navier–Stokes PDE. Lower ranks indicate stronger overall performance. DC-PINNs rank highly on violation- and stability-related metrics, with overall rank depending on how accuracy and feasibility are weighted.}
\label{fig:dc_rank_bar}
\end{figure}
Aggregates confirm the trend: DC-PINNs most consistently cut violations and oscillations while preserving accuracy. Training time rises by roughly $1.5$–$2\times$ versus PINNs, reflecting constraint enforcement, but the added cost is offset by improved stability and reliability when derivatives govern the physics.

\subsection{Scalability to Higher-Dimensional DC-PDEs}
DC-PINNs scale to higher spatial dimensions without architectural changes: as $d$ grows, the loss still combines PDE residuals with derivative equality/inequality constraints, and automatic differentiation provides all required mixed derivatives.

\paragraph{Multi-Dimensional Heat Equation.}
To illustrate this property, consider the $d$-dimensional heat equation defined on the unit hypercube $\Omega=(0,1)^d$ with zero Dirichlet boundary conditions. The governing equation in variables $\boldsymbol{x}=(x_1,\dots,x_d,t)$ is
\begin{equation}
f\!\left(\boldsymbol{x}, t\right) = \pdv{u}{t}-\lambda \sum_{i=1}^d \pdv[2]{u}{x_i}, 
\label{eq:pde_d_heat_equation}
\end{equation}
\begin{equation}
\begin{gathered}
{\mathrm s.t.}\;\; u\!\left(\boldsymbol{x}, 0\right)=\prod_{i=1}^d \sin(\pi x_i),\; u\!\left(\boldsymbol{x}, t\right)=0 \;\text{for } \boldsymbol{x}\in\partial\Omega,\\ \pdv{u}{t}\leq 0,\; \pdv[2]{u}{x_i}\leq 0 \;\text{for all } i=1,\dots,d, 
\label{eq:constraints_d_heat_equation} 
\end{gathered}
\end{equation}
The analytical solution $u(\boldsymbol{x},t)=e^{-\lambda\pi^2 d\,t}\prod_{i=1}^d \sin(\pi x_i)$ decays monotonically and satisfies the derivative signs. We set $\lambda=0.1$ and test $d\in\{1,2,8,32,128\}$; other settings follow Appendix~\ref{app:heat_impl}.

\begin{table}[htp]
\centering
\caption{Wall-clock training time (s) for the $d$-dimensional heat equation with collocation size $N_\chi$: Fixed ($N_\chi=100$), Root ($N_\chi=\lfloor100\sqrt{d}\rfloor$), Linear ($N_\chi=100d$), Power-1.5 ($N_\chi=100d^{1.5}$).}
\label{tab:high_dimensions_time}
\begin{tabular}{lcccc}
\toprule
$d$ & Fixed & Root & Linear & Power‑1.5 \\
\midrule
1   & 1.86 & 1.78 & 1.73 & 1.88 \\
2   & 1.92 & 2.06 & 2.17 & 2.44 \\
8   & 2.52 & 3.76 & 5.28 & 5.63 \\
32  & 5.76 & 6.78 & 9.58 & 30.72 \\
128 & 8.64 & 6.81 & 45.01 & --\footnote{ out‑of‑GPU-memory.} \\
\bottomrule
\end{tabular}
\end{table}

Table~\ref{tab:high_dimensions_time} shows that the training cost increases with dimension because denser sampling is required to approximate higher-dimensional solution manifolds. In this experiment, the practical limitation at large $d$ is GPU memory rather than numerical instability. However, we emphasise that this is a linear heat-equation benchmark used as a controlled stress test; these timings and stability observations do not, on their own, establish scalability for complex non-linear PDE systems.

Moving to three dimensions and beyond mainly requires two adjustments. First, the neural network must have enough capacity to represent more complex fields. Second, the collocation points must be chosen carefully so that they adequately cover the larger domain. Methods such as Latin hypercube sampling \cite{mckay2000comparison} or adaptive refinement \cite{gao2023failure, lu2021deepxde} are often effective for this purpose. Soft enforcement of constraints with adaptive weighting also avoids the numerical stiffness that often appears in hard-constraint approaches. On the computational side, DC-PINNs are well suited for data parallel training on modern accelerators like \cite{frostig2019compiling} and can incorporate techniques developed for large-scale PINNs, such as domain decomposition in XPINNs \cite{jagtap2020extended}, which splits high-dimensional domains into parallel subproblems \cite{jagtap2020extended, moseley2023finite}, and adaptive sampling.

\section{Conclusion and Future Work}
We cast the problem of solving PDEs with derivative-based equality and inequality constraints into a unified derivative-constrained optimisation framework, and proposed Derivative-Constrained Physics-Informed Neural Networks (DC-PINNs) to realise this framework. DC-PINNs combine constraint-aware losses with self-adaptive balancing to prioritise feasibility while maintaining competitive accuracy. Across the test problems considered, DC-PINNs consistently reduce violation-related metrics and stabilise training trajectories, at the cost of additional derivative evaluations and increased wall-clock time. Accuracy outcomes are problem-dependent: for Navier--Stokes, AL- and hPINNs baselines achieve lower data and residual errors under the same training budget, whereas DC-PINNs primarily improve constraint-related measures. Finally, our higher-dimensional study is based on a linear heat-equation benchmark and should be interpreted as a controlled scalability stress test rather than a proof of tractability for complex non-linear systems.

Future extensions include exploring adaptive learning rates for constraint parameters, alternative penalty formulations near inequality boundaries, and coupling with multifidelity or operator-learning architectures to further improve tractability. These directions aim to extend the applicability of DC-PINNs to broader classes of PDE-constrained systems while preserving the balance between accuracy, stability, and physical consistency.

\begin{acknowledgments}
We gratefully acknowledge anonymous reviewers for their insightful comments and constructive suggestions that have substantially improved the quality of this manuscript. We also thank the participants of the ICML 2024 AI for Science Workshop for their stimulating discussions and valuable feedback. The authors also acknowledge the Department of Computer Science at University College London for providing the high-performance computing resources that enabled this study. This work also benefited from complimentary access to cloud TPUs through Google's TPU Research Cloud (TRC) programme, whose support is sincerely acknowledged herein.
\end{acknowledgments}

\paragraph*{Code and Data Availability.}
All scripts to reproduce All Tables and FIGs are available in the repository in \\{\footnotesize \url{https://anonymous.4open.science/r/dcpinns-ef12704/}}.

\appendix
\section{The Multilayer Perceptron (MLP)}
\label{appendix:the_multi_layer_perceptron_mlp}
This appendix summarises the multilayer perceptron (MLP) used as the parametric map $u_\theta\equiv\phi_\theta$ throughout the paper and highlights how its structure supports the computation of partial derivatives with respect to inputs via automatic differentiation (AD). Let $L\ge 2$ denote the number of affine layers. The network has input dimension $d_0=n$ and output dimension $d_L=d$ (with $d\ge 1$), so that $\phi_\theta:\mathbb{R}^{n}\to\mathbb{R}^{d}$. The model is written as a composition.
\begin{equation}
    \phi_\theta(\mathbf{x})
    \;=\;
    A_L \circ f_{L-1} \circ A_{L-1} \circ \cdots \circ f_1 \circ A_1(\mathbf{x}),
    \label{eq:mlp_generative_process}
\end{equation}
where, for $l=1,\ldots,L$, each $A_l:\mathbb{R}^{d_{l-1}}\to\mathbb{R}^{d_l}$ is an affine map of the form
\begin{equation}
    A_l(\mathbf{x}_{l-1}) \;=\; W_l\,\mathbf{x}_{l-1} + \mathbf{b}_l,
    \qquad
    W_l \in \mathbb{R}^{d_l\times d_{l-1}},\;\; \mathbf{b}_l\in\mathbb{R}^{d_l},
\end{equation}
and, for $l=1,\ldots,L-1$, the activation $f_l:\mathbb{R}^{d_l}\to\mathbb{R}^{d_l}$ acts component-wise. With the conventions $\mathbf{x}_0=\mathbf{x}$ and $\mathbf{x}_l=f_l(\mathbf{z}_l)$ for $l=1,\ldots,L-1$, the forward relations are
\begin{equation}
\begin{gathered}
    \mathbf{z}_l \;=\; W_l\,\mathbf{x}_{l-1} + \mathbf{b}_l,
    \;
    \mathbf{x}_l \;=\; f_l(\mathbf{z}_l),
    \; l=1,\ldots,L-1,
    \\
    \phi_\theta(\mathbf{x}) \;=\; A_L(\mathbf{x}_{L-1}) \;=\; W_L\,\mathbf{x}_{L-1} + \mathbf{b}_L.
\end{gathered}
\end{equation}
The trainable parameters are $\theta:=\{(W_l,\mathbf{b}_l)\}_{l=1}^{L}$. Given a data set $D_0=\{(x_i^{(0)},u_i^{(0)})\}_{i=1}^{N_0}$ and a prescribed empirical risk or objective informed by physics (as defined in the main text), the parameters are obtained by minimising the loss chosen with respect to $\theta$.

For derivative-constrained formulations, derivatives of $\phi_\theta$ with respect to inputs are obtained efficiently by AD through the computational graph. Denote by $D_l(\mathbf{z}_l):=\operatorname{diag}\!\big(f_l'(\mathbf{z}_l)\big)\in\mathbb{R}^{d_l\times d_l}$ the diagonal matrix of the first derivatives of $f_l$, evaluated at $\mathbf{z}_l$. The Jacobian of $\phi_\theta$ with respect to $\mathbf{x}$ admits the compact representation.
\begin{equation}
\begin{aligned}
    J_{\phi_\theta}(\mathbf{x})
    &\;=\;
    \frac{\partial \phi_\theta}{\partial \mathbf{x}}(\mathbf{x})
    \\
    &\;=\;
    W_L \, D_{L-1}(\mathbf{z}_{L-1}) \, W_{L-1} \,\cdots\, D_1(\mathbf{z}_1)\, W_1
    \;\in\; \mathbb{R}^{d\times n},
\end{aligned}
\end{equation}
which reduces to a row vector in the scalar output case $d=1$. Higher-order derivatives, for example Hessians, can be computed via AD using compositions of Jacobian vector and vector-Jacobian products without explicitly forming dense higher-order tensors. This is standard in modern AD frameworks and suffices for the operators $\mathcal{D}$ and $\mathcal{D}_h$ used in the main text. To ensure that all derivatives appearing in the training objective are well defined, the activation functions should possess the appropriate smoothness: if derivatives of the order $k$ of $u_\theta$ enter the loss, it is sufficient to choose $f_l\in C^{k}$ for $l=1,\ldots,L-1$, for example, $\tanh$ satisfies $C^\infty$ smoothness. A formal treatment of training with input derivatives in neural networks can be found in~\cite{lo2023training}.

\section{Algorithms}
\label{section: Algorithms}
This section introduces the DC-PINNs algorithm for multi-objective problems, which controls the inequality loss of the partial derivatives of a neural network function with respect to its input features and applies the combination of loss balancing techniques for both categorised and individual losses.

\begin{center}
\begin{minipage}{0.85\linewidth}
\begin{algorithm}[H]
\DontPrintSemicolon
\caption{DC-PINNs with Balancing Processes}
\label{alg:dc-pinns}

\KwIn{Dataset $D_0$, $x_b$, $x_f$, $x_h$, $\eta$, $\eta_m$, $p_m$, $p_\lambda$, $k_{\mathrm max}$}
\KwOut{Neural network parameters $\theta$}
Consider a deep NN $\varphi_\theta(x)$ with parameters $\theta$, and the loss
\[
\mathcal{L} := \sum_\chi \lambda_{\chi}\, \hat{\mathcal{L}}_{\chi}\!\left(m_\chi, x_\chi \right),
\]
where $\hat{\mathcal{L}}_{\chi}$ denotes the categorised loss with $\chi \in \{0,b,f,h\}$, 
$m_\chi=\mathbf{1}$ are soft-weighting vectors for individual losses, and $\lambda_\chi=1$ are dynamic multipliers.\;
\For{$k \gets 1$ \KwTo $k_{\mathrm max}$}{
  Compute $\nabla_{\theta} \hat{\mathcal{L}}_\chi(k)$ by automatic differentiation.\;

  \If{$k \equiv 0 \pmod{p_m}$}{
    \vspace{8pt}
    Update $m_\chi$ by
    \[
    m_\chi^{(j)}(k+1) = m_\chi^{(j)}(k) + \eta_m \nabla_{m_\chi^{(j)}} \hat{\mathcal{L}}_\chi(k),
    \]
    where $m_\chi(k)$ and $\hat{\mathcal{L}}_\chi(k)$ denote values at the $k$-th iteration.\;
  }
  \If{$k \equiv 0 \pmod{p_\lambda}$}{
    \vspace{8pt}
    Update $\lambda_\chi$ by
    \[
    \lambda_\chi(k+1) =
    \begin{cases}
      1, & \text{if }\alpha=0,\\[4pt]
      \lambda_\chi(k) + \dfrac{\sum_\chi \alpha_\chi}{\alpha_\chi}, & \text{otherwise},
    \end{cases}
    \]
    where $\alpha_\chi = \overline{\lvert \nabla_\theta \hat{\mathcal{L}}_\chi(k)\rvert}$.
    }
  \vspace{8pt}
  Update the parameters $\theta$ via gradient descent, e.g.
  \[
  \theta(k+1) = \theta(k) - \eta \nabla_{\theta}\, \mathcal{L}(k).
  \]
}
\Return{$\theta$}
\end{algorithm}
\end{minipage}
\end{center}

The algorithm~\ref{alg:dc-pinns} exhibits DC-PINNs characteristics that separate it from conventional learning methods. First, the calculation points $x_{\left\{f,h\right\}}$ for the MLP derivatives do not correspond with the points of the training dataset $x_0$. The algorithm adjusts the derivatives to fit wide mesh grids in the defined space, thus capturing derivative data across a wide array of input features. Secondly, the objective function $\mathcal{L}$ does not depend only on the MLP's direct output but also on its derivatives as specified in Eq.\eqref{eq:categorised_losses}, all of which depend on identical network parameters. DC-PINNs facilitate the balance among the categorised losses in addition to the enhanced individual losses, which consist of PDE residuals and various scaled losses resulting from the violation of inequality constraints.

\section{Sensitivity Analysis of Adaptive Parameters}
\label{app:sensitivity_analysis_of_adaptive_parameters}
We investigate the sensitivity of DC-PINNs to the adaptive parameters $\eta_m \in \{10^{-4},10^{-3},10^{-2}\}$ and $p_m,p_\lambda \in \{10,100,1000\}$ across Heat, volatility calibration, and Navier–Stokes on dense grids with log–$\mathrm{RMSE}$.

\begin{table*}[ht]
\footnotesize
\centering
\caption{Sensitivity of DC-PINNs on the Heat problem. Reported values are $\mathrm{RMSE}$ in $\log$-scale, given as $(\mathcal{L}_0,\mathcal{L}_f,\mathcal{L}_b,\mathcal{L}_{(h1+h2)})$ in dense evaluation grids. Bold entries indicate the overall top-4 choices (by the sum of the four metrics; more negative is better).}
\label{tab:sensitivity_all_heat}
\begin{tabular}{c c|c c c}
\toprule
$\eta_m$ & $p_m \backslash p_\lambda$ & 10 & 100 & 1000 \\
\midrule
\multirow{3}{*}{$10^{-4}$} 
 & 10   & (-1.47,-2.78,-0.77,0.38) & (-1.52,-2.92,-0.83,0.30) & (-1.51,-2.89,-0.82,0.31) \\
 & 100  & (-1.40,-2.51,-0.69,0.52) & (-1.41,-2.52,-0.70,0.50) & (-1.40,-2.53,-0.69,0.51) \\
 & 1000 & (-1.33,-3.00,-0.63,0.63) & (-1.32,-2.99,-0.63,0.63) & (-1.32,-2.99,-0.62,0.63) \\
\midrule
\multirow{3}{*}{$10^{-3}$} 
 & 10   & (-4.98,-4.11,-4.50,-4.54) & (-5.20,-4.25,-4.61,-4.97) & (-5.28,-4.20,-4.69,-5.10) \\
 & 100  & (-5.12,-4.31,-4.55,-4.93) & (-5.00,-4.40,-4.47,-4.77) & (-5.23,-4.43,-4.71,-5.15) \\
 & 1000 & (-4.99,-4.15,-4.44,-4.57) & (-4.84,-4.15,-4.29,-4.33) & (-5.20,-4.36,-4.62,-5.06) \\
\midrule
\multirow{3}{*}{$10^{-2}$} 
 & 10   & (-5.61,-4.86,-5.19,-5.77) & \textbf{(-6.21,-5.25,-6.16,-7.75)} & \textbf{(-6.01,-5.10,-5.80,-6.77)} \\
 & 100  & (-5.08,-4.62,-4.45,-4.93) & \textbf{(-6.03,-5.10,-5.86,-7.11)} & \textbf{(-6.19,-5.28,-6.20,-7.97)} \\
 & 1000 & (-5.10,-4.56,-4.46,-4.89) & (-5.24,-4.54,-4.56,-5.18) & (-5.90,-5.06,-5.45,-6.60) \\
\bottomrule
\end{tabular}
\end{table*}

\begin{table*}[ht]
\footnotesize
\centering
\caption{Sensitivity of DC-PINNs on the volatility surface calibration. The reported values are $\mathrm{RMSE}$ on the logarithmic scale, given as $(\mathcal{L}0,\mathcal{L}b,\mathcal{L}f,\mathcal{L}{h{(x+xx+t)}})$ on dense evaluation grids. The bold entries indicate the overall top-4 configurations, and the lower values correspond to better performance.}
\label{tab:sensitivity_all_vol}
\begin{tabular}{c c|c c c}
\toprule
$\nu$ & $p_m \backslash p\lambda$ & 10 & 100 & 1000 \\
\midrule
\multirow{3}{*}{$10^{-4}$}
& 10 & (-4.88,-5.30,-4.16,-3.82) & (-4.86,-5.34,-4.26,-3.97) & (-4.98,-5.34,-4.33,-4.02) \\
& 100 & (-4.97,-5.03,-4.66,-4.40) & (-5.02,-4.99,-4.69,-4.41) & (-5.00,-5.05,-4.73,-4.50) \\
& 1000 & (-4.89,-4.90,-4.79,-4.50) & (-4.96,-4.94,-4.80,-4.52) & (-4.99,-4.95,-4.87,-4.51) \\
\midrule
\multirow{3}{*}{$10^{-3}$}
& 10 & (-6.76,-6.24,-6.07,-5.65) & \textbf{(-6.82,-6.40,-5.69,-5.93)} & \textbf{(-6.87,-6.35,-5.68,-5.64)} \\
& 100 & (-6.13,-5.47,-6.13,-5.05) & \textbf{(-6.57,-6.18,-6.16,-5.86)} & (-6.34,-6.00,-6.17,-5.93) \\
& 1000 & (-6.51,-6.11,-6.11,-5.71) & \textbf{(-6.59,-6.20,-6.24,-5.67)} & (-6.55,-6.10,-6.25,-5.74) \\
\midrule
\multirow{3}{*}{$10^{-2}$}
& 10 & (-1.26,-0.48,-$\infty$,-$\infty$) & (-5.27,-5.06,-4.78,-3.93) & (-6.44,-6.20,-6.26,-5.49) \\
& 100 & (-6.54,-6.09,-5.05,-4.54) & (-5.78,-5.60,-5.35,-4.95) & (-4.26,-4.29,-4.59,-3.13) \\
& 1000 & (-3.82,-3.97,-4.27,-3.20) & (-5.53,-5.45,-5.17,-4.88) & (-3.87,-4.36,-4.09,-5.90) \\
\bottomrule
\end{tabular}
\end{table*}

\begin{table*}[ht]
\footnotesize
\centering
\caption{Sensitivity of DC-PINNs on the Navier--Stokes equation. Reported values are $\mathrm{RMSE}$ in logarithmic scale, evaluated on dense grids and expressed as $(\mathcal{L}_0,\mathcal{L}_f,\mathcal{L}_h,\mu_1/\mu_2)$, where $\mathcal{L}_0=\max\{\mathcal{L}_{0_u},\mathcal{L}_{0_v},\mathcal{L}_{0_p}\}$, $\mathcal{L}_f=\max\{\mathcal{L}_{f_u},\mathcal{L}_{f_v}\}$, and $\mathcal{L}_h=\mathcal{L}_{h_{\nabla p}}$. Bold entries indicate the overall top-4 configurations, with lower values corresponding to better accuracy.}
\label{tab:sensitivity_all_ns}
{\scriptsize
\begin{tabular}{c c|c c c}
\toprule
$\nu$ & $p_m \backslash p_\lambda$ & 10 & 100 & 1000 \\
\midrule
\multirow{3}{*}{$10^{-4}$}
& 10   & (-1.00,-3.67,-1.31; 0.0483/-0.0020) & (-1.00,-3.42,-1.31; 0.0442/-0.0046) & (-1.00,-2.65,-1.31; 0.0461/-0.0084) \\
& 100  & (-1.01,-3.97,-1.31; 0.0432/0.0002) & (-1.01,-3.96,-1.31; 0.0405/0.0004) & (-1.01,-3.83,-1.31; 0.0430/-0.0008) \\
& 1000 & (-1.01,-4.28,-1.32; 0.0430/0.0012) & (-1.01,-4.27,-1.32; 0.0405/0.0013) & (-1.01,-4.12,-1.32; 0.0435/0.0010) \\
\midrule
\multirow{3}{*}{$10^{-3}$}
& 10   & (-0.94,-2.73,-1.12; 0.7695/0.0134) & (-0.94,-2.61,-1.09; 0.7109/0.0073) & (-0.94,-2.63,-1.09; 0.6992/0.0070) \\
& 100  & (-0.94,-2.79,-1.14; 0.7852/0.0160) & (-0.95,-2.78,-1.11; 0.7383/0.0095) & (-0.95,-2.98,-1.11; 0.7812/0.0117) \\
& 1000 & (-0.96,-2.99,-1.15; 0.8008/0.0165) & (-0.95,-2.90,-1.12; 0.7578/0.0118) & (-0.96,-2.92,-1.12; 0.7500/0.0099) \\
\midrule
\multirow{3}{*}{$10^{-2}$}
& 10   & (-0.93,-4.10,-1.07; 0.9805/0.0098) & (-0.93,-4.17,-1.07; 0.9805/0.0104) & (-0.92,-4.12,-1.06; 0.9844/0.0102) \\
& 100  & (-0.93,-3.99,-1.08; 0.9766/0.0115) & (-0.94,-4.00,-1.08; 0.9766/0.0099) & (-0.94,-4.12,-1.07; 0.9805/0.0104) \\
& 1000 & (-0.94,-4.14,-1.08; 0.9805/0.0112) & (-0.94,-4.16,-1.07; 0.9805/0.0106) & (-0.93,-4.06,-1.07; 0.9844/0.0103) \\
\bottomrule
\end{tabular}}
\end{table*}

On Heat case, $\eta_m$ is decisive: $10^{-4}$ underenforces constraints and destabilises the inequality term, $10^{-3}$ stabilises all losses, and $10^{-2}$ produces the lowest overall errors, especially for $\mathcal{L}_0$ and $\mathcal{L}b$. Moderate penalties ($p_m\in{10,100}$, $p\lambda\in{100,1000}$) give the most consistent gains; extremes occasionally induce stiffness and degrade derivative terms. Runtimes remain flat, so improvements reflect optimisation rather than extra computation.

In volatility calibration, the addition of $\mathcal{L}_b$ clarifies the fit–constraint trade-off, with the best joint minima concentrated at $\nu=10^{-3}$ and $p_m,p\lambda\in{100,1000}$. At $\nu=10^{-4}$ performance is stable but shallower; at $\nu=10^{-2}$ results become bimodal, with strong penalties sometimes overstiffening derivatives. For Navier–Stokes, $\mathcal{L}_0$ varies little, $\mathcal{L}_f$ improves with larger penalties (but not extreme), and $\mathcal{L}_h$ is stable; the learnt viscosity aligns with scaling, recovering $\mu_2\approx10^{-2}$.

\section{One-dimensional Heat Equation: Implementation and Experimental Details}
\label{app:heat_impl}
\subsection{Network, Training, and Sampling}
\textbf{PDE and domain.} $u_t - \lambda\,u_{xx}=0$ on $(x,t)\in[0,1]\times[0,1]$ with $u(x,0)=\sin(\pi x)$ and $u(0,t)=u(1,t)=0$; constraints of inequality $u_{xx}\le 0$ and $u_t\le 0$ in the domain.
\textbf{Network.} MLP $(x,t)\mapsto u$ with two hidden layers of width~100 and an absolute–value output activation.
\textbf{Optimiser and schedule.} The Adam optimiser is used with an exponentially decaying learning rate initialised at $10^{-3}$, where the transition step is $2000$ and the decay rate is $0.9$. The training runs for $10{,}000$ epochs in the main comparisons.
\textbf{Sampling.} Interior collocation: $31\times31$ uniform grid; validation: $101\times101$; boundaries: dense grid in $t$ (1001 points) at $x\in\{0,1\}$ and the initial line at $t=0$.
\textbf{Evaluation Metrics and Grids.} We compute log–RMSE on the initial line (\mbox{RMSE$_0$}), the interior (\mbox{RMSE$_f$}) and the boundary (\mbox{RMSE$_b$}); log–violation rates for $u_{xx}\le 0$ and $u_t\le 0$ (``$-\infty$'' indicates no violations); and log–RMSE of $u_{xx}$ and $u_t$ against the analytic solution on a $101\times101$ validation mesh. \textbf{Loss Function.} 
For the heat equation, the total loss is defined as 
$\mathcal{L} =
\lambda_{0}\hat{\mathcal{L}}_{0}
+ \lambda_{f}\hat{\mathcal{L}}_{f}
+ \lambda_{b}\hat{\mathcal{L}}_{b}
+ \lambda_{h_{\mathrm{xx}}}\hat{\mathcal{L}}_{h_{\mathrm{xx}}}
+ \lambda_{h_{\mathrm{t}}}\hat{\mathcal{L}}_{h_{\mathrm{t}}}$, 
where $\hat{\mathcal{L}}_{0}$ represents the data mismatch, 
$\hat{\mathcal{L}}_{f}$ the residual of the heat equation 
$f = \partial_t u - \lambda \partial^2_x u$, 
and $\hat{\mathcal{L}}_{b}$ the boundary loss. 
The additional components 
$\hat{\mathcal{L}}_{h_{\mathrm{x}}}$, 
$\hat{\mathcal{L}}_{h_{\mathrm{xx}}}$, 
and $\hat{\mathcal{L}}_{h_{\mathrm{t}}}$ 
measure violations of derivative-based smoothness and monotonicity conditions. 
The coefficients $\lambda$ are adaptively balanced during training to ensure stable convergence.

\section{Local Volatility Model: Methodology and Implementation Details}
\label{app:lv_method_impl}

\subsection{Methodology}
\subsubsection{Black–Scholes Pricing and Equivalence to the Heat Equation}
\label{app:lv_bs_heat}
We work on a complete filtered probability space {\small$\left(\Omega,\mathcal{F},(\mathcal{F}_t)_{t\in[0,T]},\mathbb{Q}\right)$} under the risk–neutral measure {\small$\mathbb{Q}$}. The (non-dividend-paying) asset price {\small$S_t$} follows the geometric Brownian motion
\begin{equation}
    dS_t = r\,S_t\,dt + \sigma\,S_t\,dW_t, \qquad \sigma>0,
\end{equation}
where $r$ is the risk-free rate and $W_t$ is a standard $\mathbb{Q}$-Brownian motion. The time-$t$ price of a European call with strike $K$ and maturity $T$ is
\begin{equation}
    C(S_t,K,\tau) = e^{-r\tau}\,\mathbb{E}^{\mathbb{Q}}\!\left[(S_T-K)^+\mid\mathcal{F}_t\right], \quad \tau:=T-t .
\end{equation}
By the Feynman–Kac theorem, $C$ solves the Black–Scholes (BS) PDE
\begin{equation}
\begin{gathered}
    \pdv{C}{\tau} = \tfrac{1}{2}\sigma^2 S^2 \pdv[2]{C}{S} + r S \pdv{C}{S} - r C,
    \\
    C(S,0)=(S-K)^+, \;\; \lim_{S\to0}C=0,\;\; \lim_{S\to\infty}\frac{C}{S}=1.
\end{gathered}
\end{equation}
When market quotes are expressed through \emph{implied volatility} $\sigma_{\mathrm imp}(K,\tau)$, the corresponding BS price is the closed form
\begin{equation}
\begin{gathered}
    C_{\mathrm BS}\!\left(\sigma_{\mathrm imp}\right)=S_t\,N(d_+) - K e^{-r\tau} N(d_-), \\
    d_\pm = \frac{\ln(S_t/K)+\left(r \pm \tfrac{1}{2}\sigma_{\mathrm imp}^2\right)\tau}{\sigma_{\mathrm imp}\sqrt{\tau}}, 
\end{gathered}
\label{eq: Black-Scholes formula}
\end{equation}
where $N(\cdot)$ denotes the standard normal CDF and $d_- = d_+ - \sigma_{\mathrm imp}\sqrt{\tau}$.

\paragraph{Equivalence to the Heat Equation (main text).}
The BS pricing problem is \emph{mathematically equivalent} to the one-dimensional heat equation used in the main text, cf. \eqref{eq:pde_1d_heat_equation}. In fact, set
\begin{equation}
x=\ln(S/K),\quad \theta=\tfrac{\sigma^2}{2}\tau,\quad 
\tilde{C}(x,\tau)=\frac{e^{r\tau}}{K}\,C(S,t),
\end{equation}
and define $u(x,\theta)=e^{\alpha x+\beta \theta}\tilde{C}(x,\tau)$ with
\begin{equation}
\alpha=\tfrac{1}{2}-\frac{r}{\sigma^2},\quad 
\beta=-\frac{(2r-\sigma^2)^2}{8\sigma^2}.
\end{equation}
A direct computation yields the \emph{constant-coefficient} heat equation.
\begin{equation}
    \pdv{u}{\theta}=\pdv[2]{u}{x},
\end{equation}
with the initial condition induced by the transformed payoff. Thus, the BS model and the thermodynamic heat problem are equivalent after an affine change of variables and time rescaling, resulting in the same linear parabolic PDE. In particular, the PINN/DC-PINN machinery, residual structure, and inequality handling developed for \eqref{eq:pde_1d_heat_equation} apply verbatim to the BS setting.

\subsubsection{Local Volatility as an Extension of Black–Scholes}
\label{app:lv_dupire}
The \emph{implied volatility surface} (IVS) is the map $(K,\tau)\mapsto \sigma_{\mathrm imp}(K,\tau)$ obtained by inverting the Black–Scholes formula \eqref{eq: Black-Scholes formula} on the observed prices of European options. The Local Volatility model of \cite{dupire1994pricing} extends BS by replacing the constant volatility with a \emph{deterministic function} of state and time. In the spot-time representation, one writes
\begin{equation}
dS_t = r\,S_t\,dt + \sigma_{\mathrm LV}(S_t,t)\,S_t\,dW_t,
\end{equation}
which reduces to BS when $\sigma_{\mathrm LV}(S,t)\equiv\sigma$ is constant. In the strike-maturity coordinates $(K,\tau)$, the European call prices $C(K,\tau)$ satisfy the Dupire PDE
\begin{equation}
    r\,K\,\pdv{C}{K} - \tfrac{1}{2}\,\sigma_{\mathrm LV}^2(K,\tau)\,K^2\,\pdv[2]{C}{K} + \pdv{C}{\tau} = 0,
    \label{eq: LV PDE}
\end{equation}
with boundary/initial conditions
\begin{equation}
    C(K,0)=(S_t-K)^+,\quad 
    \lim_{K\to\infty}C=0,\quad 
    \lim_{K\to0}C=S_t.
    \label{eq: boundary conditions of PDE}
\end{equation}
Hence, relative to BS, LV is (i) a state- and time-inhomogeneous diffusion, which is spatially varying diffusivity in the heat-equation analogy, and (ii) rich enough to fit the entire IVS while remaining Markov and arbitrage-free at the level of European options.

Practically, one often calibrates $\sigma_{\mathrm LV}$ from a sufficiently smooth $\sigma_{\mathrm imp}(K,\tau)$ by substituting BS prices in \eqref{eq: Black-Scholes formula} into Dupire’s equation \eqref{eq: LV PDE}, which yields the conversion
\begin{equation}
\begin{aligned}
    &\sigma_{\mathrm{LV}}^2(K, \tau)\\
    &= \tfrac{\sigma_{\mathrm{imp}}^2
        + 2 \sigma_{\mathrm{imp}}\tau\!\left(\pdv{\sigma_{\mathrm{imp}}}{\tau}
        + r K \pdv{\sigma_{\mathrm{imp}}}{K}\right)}
    {\,1 + 2 d_+ K \sqrt{\tau}\,\pdv{\sigma_{\mathrm{imp}}}{K}
        + K^2 \tau\!\left(d_+ d_- \!\left(\pdv{\sigma_{\mathrm{imp}}}{K}\right)^{\!2}
        + \sigma_{\mathrm{imp}} \pdv[2]{\sigma_{\mathrm{imp}}}{K}\right)}.
\label{eq:LV-vol-from-BS-vol}
\end{aligned}
\end{equation}
where $d_\pm$ are the BS quantities defined below \eqref{eq: Black-Scholes formula}. Equation \eqref{eq:LV-vol-from-BS-vol} makes explicit that LV is an \emph{extension} of BS: if $\sigma_{\mathrm imp}$ is constant in $(K,\tau)$, then $\sigma_{\mathrm LV}\equiv\sigma_{\mathrm imp}$ and \eqref{eq: LV PDE} collapses in the BS case.

\subsubsection{No-Arbitrage Constraints for European Options}
\label{app:lv_no_arb}
The option prices should obey the restrictions imposed by no-arbitrage conditions, which are essential financial principles that posit that market prices prevent guaranteed returns above the risk-free rate. This study considers the necessary and sufficient conditions for no-arbitrage presented in \cite{carr2005note}. This allows us to express the call option price as a two-dimensional surface appropriately. The necessary and sufficient conditions for no-arbitrage are represented as non-strict inequalities for several first and second derivatives,
\begin{equation}
\begin{gathered}
    -e^{-r\tau} \leq \pdv{C}{K}\leq 0,\;\;\; \pdv[2]{C}{K} \geq 0,\;\;\; \pdv{C}{\tau}\geq 0.
    \label{eq: no-arbitrage conditions}
\end{gathered}    
\end{equation}
From the above, the no-arbitrage conditions require these derivatives to have a specific sign. The standard architecture does not automatically satisfy these conditions when calibrating with a loss function simply based on the mean squared error (MSE) of the prices.

\subsection{Network, Training, and Sampling}
\label{app:lv_impl:training}
\textbf{Problem and domain.} We learn a price surface $C(K,T)$ under a synthetic local volatility model with spot $s_0=1$, rate $r=0.05$ and maturity $T\in(0,1]$, enforcing monotonicity/convexity in $K$ and temporal smoothness through $h_x$, $h_{xx}$ and $h_t$.
\textbf{Interior and validation grids.} Interior collocation uses a $50\times50$ uniform $(K,T)$ grid with $K\in[0.1,2.0]$ and $T\in[0.01,1.0]$; validation uses a $100\times100$ grid over the same ranges.
\textbf{Boundary/initial observations.} The data set provides price observations on (i) the $T{=}0$ line and (ii) selected boundary slices, which are used as supervised anchors alongside the PDE residual.
\textbf{Network.} An MLP with four hidden layers of width~50 and a Softplus output activation maps $(K,T)\mapsto C$.
\textbf{Optimiser and schedule.} Adam optimiser is used with an exponentially decaying learning rate initialised at $10^{-3}$, where the transition step is $2000$ and the decay rate is $0.9$. The training runs for $10{,}000$ epochs in the main comparisons.
\textbf{Sampling.} 
The spot follows the diffusion of local volatility under the risk-neutral measure
$\mathrm{d}S_t = rS_t\,\mathrm{d}t + \sigma_{\mathrm LV}(S_t,t)\,S_t\,\mathrm{d}W_t,$ 
\begin{equation}
\sigma_{\mathrm LV}(S,t) = \sigma_A + \tfrac{A}{S} + Bt,
\end{equation}
where the parameters are fixed as $\sigma_A=0.2$, $A=0.1$, and $B=0.2$, following the setup in \citep{kim2021reconstruction,boyle2000volatility}. The simulation domain is $S\in[0.1S_0,2S_0]$ and $t\in[0,1]$. European call prices are computed by Monte Carlo using a log Euler discretisation with uniform time steps $N_{\text{steps}}$ $\Delta t=T/N_{\text{steps}}$. The update rule
\begin{gather*}
S_{t+\Delta t}=S_t\exp\!\left((r-q-\tfrac{1}{2}\sigma^2)\Delta t+\sigma\sqrt{\Delta t}\,Z\right),\\
Z\sim\mathcal N(0,1),
\end{gather*}
preserves positivity and evaluates $\sigma_{\mathrm LV}(S_t,t)$ pathwise at each step. Antithetic sampling is applied by pairing $Z$ with $-Z$ to reduce variance. The discounted Monte Carlo estimator for a call with strike $K$ is
\[
C^{\text{MC}}(S_0,K,T)=e^{-rT}\,\frac{1}{N_{\text{paths}}}\sum_{i=1}^{N_{\text{paths}}}\max\!\bigl(S_T^{(i)}-K,0\bigr).
\]
Prices are generated on a grid of maturities and logarithmic moneyness values, with $T\in[0,1]$ and $y=\log(K/S_0)\in[-2,2]$. We set $S_0=1$ and $r=0.05$. The scheme is first-order accurate in $\Delta t$, with Monte Carlo standard error $\mathcal{O}(N_{\text{paths}}^{-1/2})$. The lower bound $S\ge0.1S_0$ keeps the $A/S$ term finite, while antithetic sampling improves stability for short maturities.
The computed call prices $u_{i, j}$ were then polluted by a relative noise $u^{\text {noisy }}=u\left(1+0.05 \eta\right)$ with $\eta$ drawn from the standard normal distribution $\mathcal{N}(0,1)$. Observed prices are sampled as $x\sim \mathrm{TruncNorm}([-1,1])$ and $t\sim \mathrm{TruncNorm}([0.1,2.0])$, thus covering the relevant ranges of moneyness and maturity. We use $N_0=1{,}000$ observed option prices, a $51\times 51$ collocation grid for the interior residual, and $N_h=5{,}000$ inequality-enforcement points. \textbf{Loss Function.} The total loss is defined as 
$\mathcal{L} =
\lambda_{0}\hat{\mathcal{L}}_{0}
+ \lambda_{f}\hat{\mathcal{L}}_{f}
+ \lambda_{b}\hat{\mathcal{L}}_{b}
+ \lambda_{h_{\mathrm{x}}}\hat{\mathcal{L}}_{h_{\mathrm{x}}}
+ \lambda_{h_{\mathrm{xx}}}\hat{\mathcal{L}}_{h_{\mathrm{xx}}}
+ \lambda_{h_{\mathrm{t}}}\hat{\mathcal{L}}_{h_{\mathrm{t}}}$, 
where $\hat{\mathcal{L}}_{0}$ denotes the data-mismatch term, $\hat{\mathcal{L}}_{f}$ the PDE residual, and $\hat{\mathcal{L}}_{b}$ the boundary loss. 
The additional terms $\hat{\mathcal{L}}_{h_{\mathrm{x}}}$, $\hat{\mathcal{L}}_{h_{\mathrm{xx}}}$, and $\hat{\mathcal{L}}_{h_{\mathrm{t}}}$ quantify violations of the derivative-based no-arbitrage conditions. 
The coefficients $\lambda$ are adaptively balanced during training to stabilise the optimisation dynamics.

\section{Incompressible Flow Dynamics: Methodology and Implementation Details}
\label{app:ns_method_impl}
\subsection{Reference Simulation and Data Extraction}
High-fidelity reference data were obtained using the spectral/hp element method implemented in \textsc{Nektar++} \citep{moxey2020nektar++}. An unstructured triangular mesh refined around the cylinder resolves the near-wall shear and the downstream vortex street. The governing equations are the two-dimensional incompressible Navier--Stokes equations in a velocity--pressure formulation, advanced by the velocity-correction (i.e. projection) scheme with a Galerkin projection. Advection was discretised in convective form and diffusion was treated implicitly. Time progression was used with a second order IMEX scheme with a fixed step $\Delta t = 0.025$.

Unless otherwise stated, the nominal Reynolds number was $\mathrm{Re}=100$ (kinematic viscosity $\nu = 1/\mathrm{Re}$), with spectral/hp dealiasing enabled to control aliasing errors introduced by non-linear advection. To further attenuate spurious energy in the highest-resolved modes, spectral vanishing viscosity with a DG kernel and cutoff ratio $0.5$ was applied. Global systems arising from the pressure Poisson and elliptic subproblems were solved using direct static condensation. Boundary conditions follow the standard flow cylinder configuration: no-slip velocity on the surface of the cylinder (wall), a uniform inflow $(u,v)=(1,0)$ at the inlet, traction-free conditions at the outlet and weak Dirichlet far field conditions elsewhere. Polynomial expansions used for $(u,v,p)$.

The calculation was integrated until a statistically steady periodic shedding regime was established at $t \approx 800$. After discarding the initial transient, flow field snapshots $(u,v,p)$ were sampled in every $0.1$ time unit to form the reference data set. In addition to volumetric fields, we recorded the aerodynamic force coefficients on the cylinder wall, a probe time series at a fixed monitoring location in the wake, and per-mode kinetic energies. These diagnostics were used to verify convergence to the limit cycle (through the lift coefficient and probe spectra) and to compute the Strouhal number.

All volumetric outputs were written in VTU format and post-processed in ParaView \cite{ayachit2015paraview}. A reproducible Python script accompanying the data colours the scene by the stream-wise velocity, applies an arrow-glyph overlay for velocity vectors, and stores camera state and screenshots used for figures in this paper.

\subsection{Network, Training, and Sampling}
\label{app:ns_impl:training}
\textbf{Problem.} Two-dimensional incompressible flow past a circular cylinder; we use the precomputed wake dataset provided with the codebase. The computational domain and boundary conditions are as stated in the main text.
\textbf{Network.} An eight–layer MLP of width~20 maps $(x,y,t)\mapsto(\phi,p)$; inputs are linearly rescaled to $[-1,1]$ before the final layer. Parameters are initialised by Glorot initialisation.
\textbf{Optimiser and schedule.} Adam optimiser is used with an exponentially decaying learning rate initialised at $10^{-3}$, where the transition step is $2000$ and the decay rate is $0.9$. The training runs for $10{,}000$ epochs in the main comparisons.
\textbf{Physics and constraints.} To preserve physical admissibility, we enforce incompressibility, $\nabla\cdot\mathbf{V}=0$, and bound the pressure gradient by $|\nabla p|\le\tfrac{1}{2}\rho U^2$, which limits excessive shear and ensures numerically stable flow fields, where $\rho=1$ and $U=1$ in the dimensionless framework based on the diameter of the cylinder and the flow speed. The viscosity coefficients $(\mu_1,\mu_2)$ are learnt together with the network; throughout the sweeps, $\mu_2$ converged to $\approx0.01$.
\textbf{Regularisation.} Let $U$ denote the reference velocity scales, as non-dimensionalisation, with diameter $D$ and inflow $U_\infty$, $\tilde{x}=x/D$, $\tilde{t}=t\,U_\infty/D$, $(u,v)=U_\infty(\tilde{u},\tilde{v})$, and viscosity $\nu$, so that $\mu_1=1$ and $\mu_2=\nu/(U_\infty D)=1/\mathrm{Re}$. All losses and constraints are computed in non-dimensional variables. \textbf{Sampling.} From these simulations, we extract velocity samples $5{,}000$ in the near-wake grids ($x\in[1,8],y\in[-2,2]$) and the location grids for PDE residuals and constraints, as follows \cite{raissi2019physics}. \textbf{Loss Function.} The total loss is $\mathcal{L} :=
\lambda_{0_u} \hat{\mathcal{L}}_{0_u}
+ \lambda_{0_v} \hat{\mathcal{L}}_{0_v}
+ \lambda_{f^u} \hat{\mathcal{L}}_{f^u}
+ \lambda_{f^v} \hat{\mathcal{L}}_{f^v}
+ \lambda_{h_{\mathrm{\nabla p}}}\hat{\mathcal{L}}_{h_{\mathrm{\nabla p}}}$,
where $\hat{\mathcal{L}}_{0_u/0_v}$ are the data mismatch, $\hat{\mathcal{L}}_{f_u/f_v}$ are PDE residuals, $\hat{\mathcal{L}_{h^{\mathrm{\nabla p}}}}$ enforces the upper bound following Bernoulli's principle. The coefficients $\lambda$ are adaptively balanced during training.

\subsection{Comparative Analysis}
We compare DC-PINNs with standard PINNs and the Nektar++ reference FEM solver. The metrics include RMSE of $(u,v,p)$, divergence error, violation rates of inequality constraints, and running time efficiency. The results show that DC-PINNs maintain incompressibility, reduce shear violations, and achieve pressure/velocity errors comparable to those of FEM while outperforming baseline PINNs in stability and physical consistency.

\begin{figure}[htbp]
    \includegraphics[width=\columnwidth]{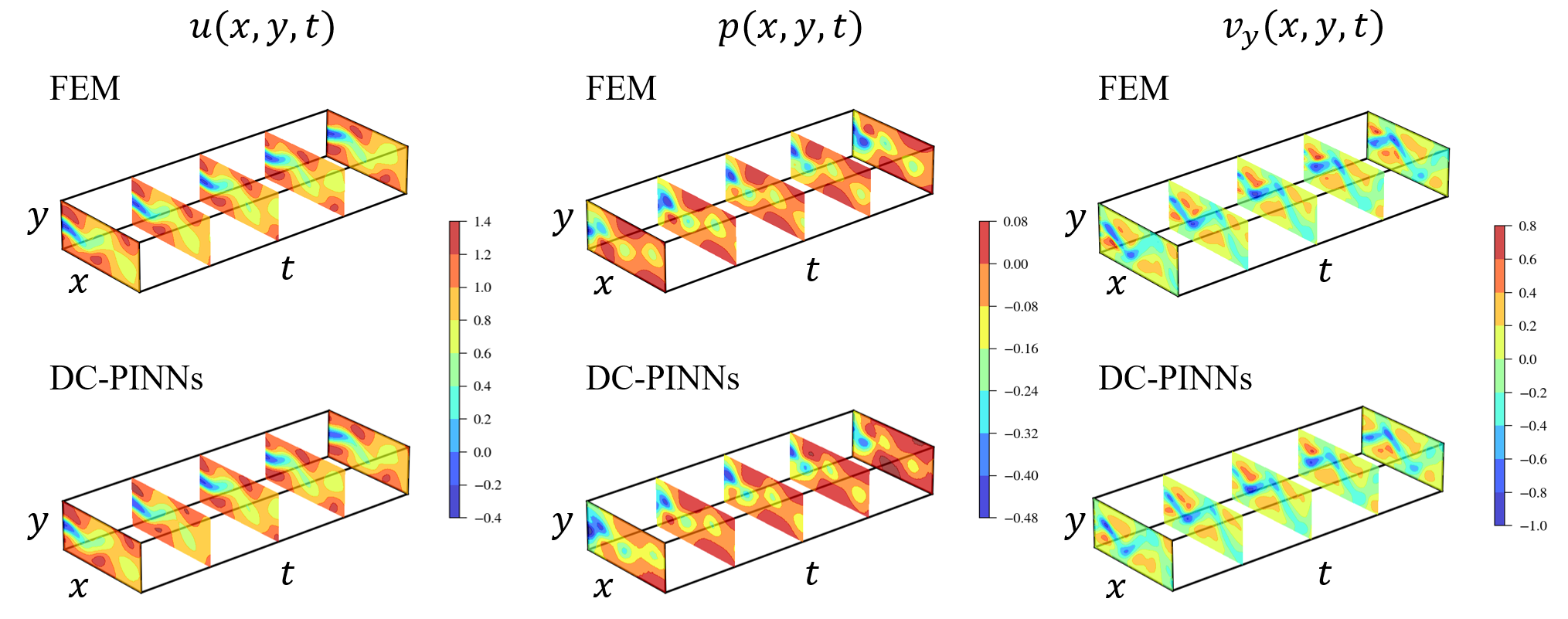}
    \caption{Flow past a circular cylinder at $Re=100$. Shown are the spatio–temporal fields $u(x,y,t)$, $p(x,y,t)$ (average adjusted), and $v_y(x,y,t)$. The top row provides the FEM reference and the bottom row shows DC-PINNs. The slabs indicate several time instants along the $t$-axis.}
    \label{fig:DC-PINNs_NavierStokes_movie}
\end{figure}

The results are compared with those obtained using a spectral/hp element method implemented in the open source software Nektar++ \citep{moxey2020nektar++}, as illustrated in FIG. \ref{fig:DC-PINNs_NavierStokes_movie}. In training, velocity and pressure fields are defined $\left(\phi, p\right) := \psi_\theta\left(x, y, t\right), \left(u, v\right)= \left({\partial \phi}/{\partial y}, -{\partial \phi}/{\partial x}\right)$, following the architecture in \cite{raissi2019physics}. In the experiment, we consider the divergence-free field and vortices shed in the wake having a characteristic size comparable to the diameter of the cylinder as derivative constraints, inspired by \cite{singh2005vortex}, in \eqref{eq: constraints Navier Stokes}.

In FIG. \ref{fig:DC-PINNs_NavierStokes_movie}, DC-PINNs closely reproduces the FEM reference across all three fields. The large-scale vortex-shedding patterns align in phase and wavelength in $u$, the pressure field $p$ exhibits comparable low-amplitude structures without spurious oscillations, and the cross-stream velocity $v_y$ matches the alternating streaks in the wake. Relative to FEM, DC-PINNs shows mild smoothing of sharp extrema near the shear layers, yet the coherent structures and their temporal evolution remain consistent, indicating a faithful reconstruction of the wake dynamics on the evaluation grids.

\begin{figure}[htbp]
    \includegraphics[width=\columnwidth]{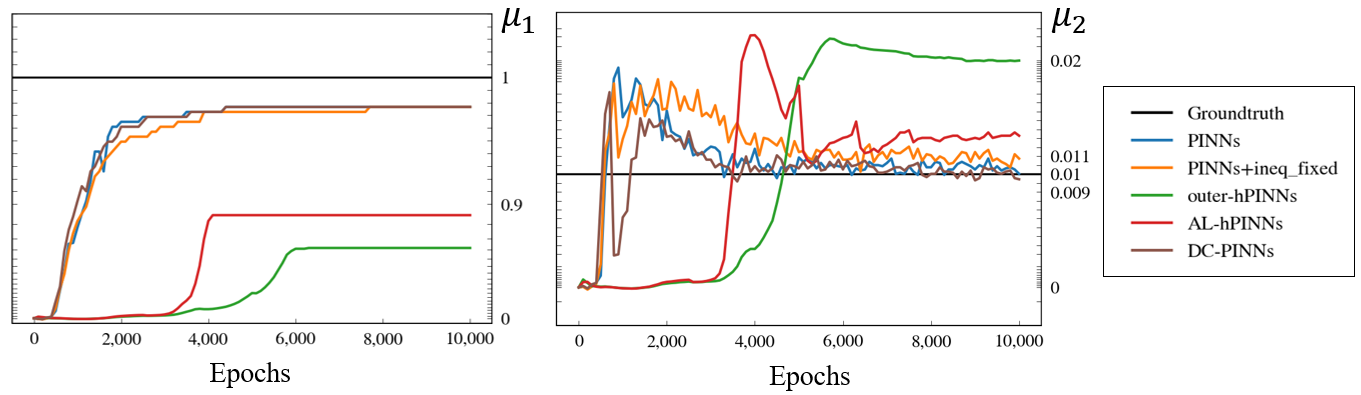}
    \caption{Evolution of trained coefficients ($\mu_{1/2}$) by epochs.}
    \label{fig:DC-PINNs_NS_mus}
\end{figure}

To evaluate the performance of DC-PINNs, FIG. \ref{fig:DC-PINNs_NS_mus} presents the learning of system identification, or data-driven discovery of PDEs, showing how the parameters $\mu = (\mu_1, \mu_2)$ describe the observed data. DC-PINNs effectively converge to the reference values for the parameters $\mu$ during the training process. The results show that DC-PINNs can accurately capture complex flow features, including the von Kármán vortex street in the cylinder wake. The additional physical constraints enforced by DC-PINNs help maintain solution stability and prevent nonphysical artefacts that may arise in unconstrained neural network approaches.

\bibliographystyle{apsrev4-2}
\bibliography{references}

@article{raissi2019physics,
  title={Physics-informed neural networks: A deep learning framework for solving forward and inverse problems involving nonlinear partial differential equations},
  author={Raissi, Maziar and Perdikaris, Paris and Karniadakis, George E},
  journal={Journal of Computational physics},
  volume={378},
  pages={686--707},
  year={2019},
  publisher={Elsevier}
}

@article{wang2023expert,
  title={An expert's guide to training physics-informed neural networks},
  author={Wang, Sifan and Sankaran, Shyam and Wang, Hanwen and Perdikaris, Paris},
  journal={arXiv preprint arXiv:2308.08468},
  year={2023}
}

@article{dupire1994pricing,
  title={Pricing with a smile},
  author={Dupire, Bruno and others},
  journal={Risk},
  volume={7},
  number={1},
  pages={18--20},
  year={1994}
}

@article{carr2005note,
  title={A note on sufficient conditions for no arbitrage},
  author={Carr, Peter and Madan, Dilip B},
  journal={Finance Research Letters},
  volume={2},
  number={3},
  pages={125--130},
  year={2005},
  publisher={Elsevier}
}

@article{wang2022respecting,
  title={Respecting causality is all you need for training physics-informed neural networks},
  author={Wang, Sifan and Sankaran, Shyam and Perdikaris, Paris},
  journal={arXiv preprint arXiv:2203.07404},
  year={2022}
}

@misc{jax2018github,
  author = {James Bradbury and Roy Frostig and Peter Hawkins and Matthew James Johnson and Chris Leary and Dougal Maclaurin and George Necula and Adam Paszke and Jake Vander{P}las and Skye Wanderman-{M}ilne and Qiao Zhang},
  title = {{JAX}: composable transformations of {P}ython+{N}um{P}y programs},
  url = {http://github.com/google/jax},
  version = {0.3.13},
  year = {2018},
}

@misc{flax2020github,
  author = {Jonathan Heek and Anselm Levskaya and Avital Oliver and Marvin Ritter and Bertrand Rondepierre and Andreas Steiner and Marc van {Z}ee},
  title = {{F}lax: A neural network library and ecosystem for {JAX}},
  url = {http://github.com/google/flax},
  version = {0.8.2},
  year = {2023},
}

@article{kingma2014adam,
  title={Adam: A method for stochastic optimization},
  author={Kingma, Diederik P and Ba, Jimmy},
  journal={arXiv preprint arXiv:1412.6980},
  year={2014}
}

@article{jagtap2020conservative,
  title={Conservative physics-informed neural networks on discrete domains for conservation laws: Applications to forward and inverse problems},
  author={Jagtap, Ameya D and Kharazmi, Ehsan and Karniadakis, George Em},
  journal={Computer Methods in Applied Mechanics and Engineering},
  volume={365},
  pages={113028},
  year={2020},
  publisher={Elsevier}
}

@article{lagaris1998artificial,
  title={Artificial neural networks for solving ordinary and partial differential equations},
  author={Lagaris, Isaac E and Likas, Aristidis and Fotiadis, Dimitrios I},
  journal={IEEE transactions on neural networks},
  volume={9},
  number={5},
  pages={987--1000},
  year={1998},
  publisher={IEEE}
}

@article{lu2021physics,
  title={Physics-informed neural networks with hard constraints for inverse design},
  author={Lu, Lu and Pestourie, Raphael and Yao, Wenjie and Wang, Zhicheng and Verdugo, Francesc and Johnson, Steven G},
  journal={SIAM Journal on Scientific Computing},
  volume={43},
  number={6},
  pages={B1105--B1132},
  year={2021},
  publisher={SIAM}
}

@book{cannon1984one,
  title={The one-dimensional heat equation},
  author={Cannon, John Rozier},
  number={23},
  year={1984},
  publisher={Cambridge University Press}
}

@article{chen2021theory,
  title={Theory-guided hard constraint projection (HCP): A knowledge-based data-driven scientific machine learning method},
  author={Chen, Yuntian and Huang, Dou and Zhang, Dongxiao and Zeng, Junsheng and Wang, Nanzhe and Zhang, Haoran and Yan, Jinyue},
  journal={Journal of Computational Physics},
  volume={445},
  pages={110624},
  year={2021},
  publisher={Elsevier}
}

@article{lo2023training,
  title={On Training Derivative-Constrained Neural Networks},
  author={Lo, KaiChieh and Huang, Daniel},
  journal={arXiv preprint arXiv:2310.01649},
  year={2023}
}

@article{moxey2020nektar++,
  title={Nektar++: Enhancing the capability and application of high-fidelity spectral/hp element methods},
  author={Moxey, David and Cantwell, Chris D and Bao, Yan and Cassinelli, Andrea and Castiglioni, Giacomo and Chun, Sehun and Juda, Emilia and Kazemi, Ehsan and Lackhove, Kilian and Marcon, Julian and others},
  journal={Computer Physics Communications},
  volume={249},
  pages={107110},
  year={2020},
  publisher={Elsevier}
}

@article{singh2005vortex,
  title={Vortex-induced oscillations at low Reynolds numbers: hysteresis and vortex-shedding modes},
  author={Singh, SP and Mittal, S},
  journal={Journal of fluids and structures},
  volume={20},
  number={8},
  pages={1085--1104},
  year={2005},
  publisher={Elsevier}
}

@article{mcclenny2023self,
  title={Self-adaptive physics-informed neural networks},
  author={McClenny, Levi D and Braga-Neto, Ulisses M},
  journal={Journal of Computational Physics},
  volume={474},
  pages={111722},
  year={2023},
  publisher={Elsevier}
}

@article{son2023enhanced,
  title={Enhanced physics-informed neural networks with augmented Lagrangian relaxation method (AL-PINNs)},
  author={Son, Hwijae and Cho, Sung Woong and Hwang, Hyung Ju},
  journal={Neurocomputing},
  volume={548},
  pages={126424},
  year={2023},
  publisher={Elsevier}
}

@article{patel2024turbulence,
  title={Turbulence model augmented physics-informed neural networks for mean-flow reconstruction},
  author={Patel, Yusuf and Mons, Vincent and Marquet, Olivier and Rigas, Georgios},
  journal={Physical Review Fluids},
  volume={9},
  number={3},
  pages={034605},
  year={2024},
  publisher={APS}
}

@article{karniadakis2021physics,
  title={Physics-informed machine learning},
  author={Karniadakis, George Em and Kevrekidis, Ioannis G and Lu, Lu and Perdikaris, Paris and Wang, Sifan and Yang, Liu},
  journal={Nature Reviews Physics},
  volume={3},
  number={6},
  pages={422--440},
  year={2021},
  publisher={Nature Publishing Group UK London}
}

@article{chen2021physics,
  title={Physics-informed learning of governing equations from scarce data},
  author={Chen, Zhao and Liu, Yang and Sun, Hao},
  journal={Nature communications},
  volume={12},
  number={1},
  pages={6136},
  year={2021},
  publisher={Nature Publishing Group UK London}
}

@article{sirignano2018dgm,
  title={DGM: A deep learning algorithm for solving partial differential equations},
  author={Sirignano, Justin and Spiliopoulos, Konstantinos},
  journal={Journal of computational physics},
  volume={375},
  pages={1339--1364},
  year={2018},
  publisher={Elsevier}
}

@article{berg2018unified,
  title={A unified deep artificial neural network approach to partial differential equations in complex geometries},
  author={Berg, Jens and Nystr{\"o}m, Kaj},
  journal={Neurocomputing},
  volume={317},
  pages={28--41},
  year={2018},
  publisher={Elsevier}
}

@article{yu2018deep,
  title={The deep Ritz method: a deep learning-based numerical algorithm for solving variational problems},
  author={Yu, Bing and others},
  journal={Communications in Mathematics and Statistics},
  volume={6},
  number={1},
  pages={1--12},
  year={2018},
  publisher={Springer}
}

@article{kharazmi2021hp,
  title={hp-VPINNs: Variational physics-informed neural networks with domain decomposition},
  author={Kharazmi, Ehsan and Zhang, Zhongqiang and Karniadakis, George Em},
  journal={Computer Methods in Applied Mechanics and Engineering},
  volume={374},
  pages={113547},
  year={2021},
  publisher={Elsevier}
}

@article{onsager1931reciprocal,
  title={Reciprocal relations in irreversible processes. I.},
  author={Onsager, Lars},
  journal={Physical review},
  volume={37},
  number={4},
  pages={405},
  year={1931},
  publisher={APS}
}

@article{sukumar2022exact,
  title={Exact imposition of boundary conditions with distance functions in physics-informed deep neural networks},
  author={Sukumar, Natarajan and Srivastava, Ankit},
  journal={Computer Methods in Applied Mechanics and Engineering},
  volume={389},
  pages={114333},
  year={2022},
  publisher={Elsevier}
}

@article{beucler2021enforcing,
  title={Enforcing analytic constraints in neural networks emulating physical systems},
  author={Beucler, Tom and Pritchard, Michael and Rasp, Stephan and Ott, Jordan and Baldi, Pierre and Gentine, Pierre},
  journal={Physical review letters},
  volume={126},
  number={9},
  pages={098302},
  year={2021},
  publisher={APS}
}

@article{wang2022and,
  title={When and why PINNs fail to train: A neural tangent kernel perspective},
  author={Wang, Sifan and Yu, Xinling and Perdikaris, Paris},
  journal={Journal of Computational Physics},
  volume={449},
  pages={110768},
  year={2022},
  publisher={Elsevier}
}

@book{nocedal2006numerical,
  title={Numerical optimization},
  author={Nocedal, Jorge and Wright, Stephen J},
  year={2006},
  publisher={Springer}
}

@book{bertsekas2014constrained,
  title={Constrained optimization and Lagrange multiplier methods},
  author={Bertsekas, Dimitri P},
  year={2014},
  publisher={Academic press}
}

@book{peyret2012computational,
  title={Computational methods for fluid flow},
  author={Peyret, Roger and Taylor, Thomas D},
  year={2012},
  publisher={Springer Science \& Business Media}
}

@book{batchelor2000introduction,
  title={An introduction to fluid dynamics},
  author={Batchelor, George Keith},
  year={2000},
  publisher={Cambridge university press}
}

@article{googlecolab,
  title={Frequently Asked Questions. Available online:},
  author={Google},
  journal={https://research.google.com/colaboratory/faq.html (accessed on 25th Sep 2025)},
  year={2025}
}

@inproceedings{glorot2010understanding,
  title={Understanding the difficulty of training deep feedforward neural networks},
  author={Glorot, Xavier and Bengio, Yoshua},
  booktitle={Proceedings of the thirteenth international conference on artificial intelligence and statistics},
  pages={249--256},
  year={2010},
  organization={JMLR Workshop and Conference Proceedings}
}

@inproceedings{frostig2019compiling,
  title={Compiling machine learning programs via high-level tracing},
  author={Frostig, Roy and Johnson, Matthew James and Leary, Chris},
  booktitle={SysML conference 2018},
  year={2019}
}

@article{jagtap2020extended,
  title={Extended physics-informed neural networks (XPINNs): A generalized space-time domain decomposition based deep learning framework for nonlinear partial differential equations},
  author={Jagtap, Ameya D and Karniadakis, George Em},
  journal={Communications in Computational Physics},
  volume={28},
  number={5},
  year={2020},
  publisher={Brown Univ., Providence, RI (United States)}
}

@article{moseley2023finite,
  title={Finite basis physics-informed neural networks (FBPINNs): a scalable domain decomposition approach for solving differential equations},
  author={Moseley, Ben and Markham, Andrew and Nissen-Meyer, Tarje},
  journal={Advances in Computational Mathematics},
  volume={49},
  number={4},
  pages={62},
  year={2023},
  publisher={Springer}
}

@article{mckay2000comparison,
  title={A comparison of three methods for selecting values of input variables in the analysis of output from a computer code},
  author={McKay, Michael D and Beckman, Richard J and Conover, William J},
  journal={Technometrics},
  volume={42},
  number={1},
  pages={55--61},
  year={2000},
  publisher={Taylor \& Francis}
}

@article{gao2023failure,
  title={Failure-informed adaptive sampling for PINNs},
  author={Gao, Zhiwei and Yan, Liang and Zhou, Tao},
  journal={SIAM Journal on Scientific Computing},
  volume={45},
  number={4},
  pages={A1971--A1994},
  year={2023},
  publisher={SIAM}
}

@article{lu2021deepxde,
  title={DeepXDE: A deep learning library for solving differential equations},
  author={Lu, Lu and Meng, Xuhui and Mao, Zhiping and Karniadakis, George Em},
  journal={SIAM review},
  volume={63},
  number={1},
  pages={208--228},
  year={2021},
  publisher={SIAM}
}

@article{kim2021reconstruction,
  title={Reconstruction of the local volatility function using the Black--Scholes model},
  author={Kim, Sangkwon and Han, Hyunsoo and Jang, Hanbyeol and Jeong, Darae and Lee, Chaeyoung and Lee, Wonjin and Kim, Junseok},
  journal={Journal of Computational Science},
  volume={51},
  pages={101341},
  year={2021},
  publisher={Elsevier}
}

@article{boyle2000volatility,
  title={Volatility estimation from observed option prices},
  author={Boyle, Phelim P and Thangaraj, Draviam},
  journal={Decisions in Economics and Finance},
  volume={23},
  number={1},
  pages={31--52},
  year={2000},
  publisher={Springer}
}

@book{ayachit2015paraview,
  title={The paraview guide: a parallel visualization application},
  author={Ayachit, Utkarsh},
  year={2015},
  publisher={Kitware, Inc.}
}

@article{krishnapriyan2021characterizing,
  title={Characterizing possible failure modes in physics-informed neural networks},
  author={Krishnapriyan, Aditi and Gholami, Amir and Zhe, Shandian and Kirby, Robert and Mahoney, Michael W},
  journal={Advances in neural information processing systems},
  volume={34},
  pages={26548--26560},
  year={2021}
}

@article{cuomo2022scientific,
  title={Scientific machine learning through physics--informed neural networks: Where we are and what’s next},
  author={Cuomo, Salvatore and Di Cola, Vincenzo Schiano and Giampaolo, Fabio and Rozza, Gianluigi and Raissi, Maziar and Piccialli, Francesco},
  journal={Journal of Scientific Computing},
  volume={92},
  number={3},
  pages={88},
  year={2022},
  publisher={Springer}
}

\end{document}